\pdfoutput=1

\documentclass[11pt]{article}

\usepackage[]{EMNLP2022}

\usepackage{times}
\usepackage{latexsym}

\usepackage[T1]{fontenc}

\usepackage[utf8]{inputenc}

\usepackage{microtype}

\usepackage{inconsolata}
\usepackage{amsmath}
\usepackage{amssymb}
\usepackage{enumitem}
\usepackage{booktabs}
\usepackage{multicol}
\usepackage{multirow}
\usepackage{float}
\usepackage{graphics}
\usepackage{graphicx}
\usepackage{makecell}
\usepackage{array}
\usepackage{bm}
\usepackage{booktabs}
\usepackage{hyperref}

\setenumerate[1]{itemsep=0pt,partopsep=0pt,parsep=\parskip,topsep=5pt}
\setitemize[1]{itemsep=0pt,partopsep=0pt,parsep=\parskip,topsep=5pt}
\newcolumntype{L}[1]{>{\raggedright\let\newline\\\arraybackslash\hspace{0pt}}m{#1}}
\newcolumntype{C}[1]{>{\centering\let\newline\\\arraybackslash\hspace{0pt}}m{#1}}
\newcolumntype{R}[1]{>{\raggedleft\let\newline\\\arraybackslash\hspace{0pt}}m{#1}}
\setenumerate[1]{itemsep=0pt,partopsep=0pt,parsep=\parskip,topsep=5pt}
\setitemize[1]{itemsep=0pt,partopsep=0pt,parsep=\parskip,topsep=5pt}
\usepackage{wasysym}
\usepackage{upgreek}
\usepackage{color}

\title{Prompt Conditioned VAE: Enhancing Generative Replay \\for Lifelong Learning in Task-Oriented Dialogue}

\author{
Yingxiu Zhao$^{1}$\thanks{\quad Work done while the author was interning at Alibaba.}, \
Yinhe Zheng$^2$\thanks{\quad \small Corresponding author.}, \ 
Zhiliang Tian$^1$, \ 
Chang Gao$^3$, \
Bowen Yu$^2$, \\
\textbf{
Haiyang Yu$^2$, \
Yongbin Li$^2$, \
Jian Sun$^2$, \ 
Nevin L. Zhang$^1$} \\
$^1$ The Hong Kong University of Science and Technology\\
$^2$ Alibaba Group,  
$^3$ The Chinese University of Hong Kong \\
\small \texttt{\{yzhaocx,ztianac,lzhang\}@connect.ust.hk}, \texttt{zhengyinhe1@163.com},\\ 
\small \quad \texttt{gaochang@se.cuhk.edu.hk, jiansun\_china@hotmail.com}\\
\small \quad \texttt{\{yubowen.ybw,yifei.yhy,shuide.lyb\}@alibaba-inc.com}
}

\begin{document}
\maketitle
\begin{abstract}
Lifelong learning (LL) is vital for advanced task-oriented dialogue (ToD) systems. To address the catastrophic forgetting issue of LL, generative replay methods are widely employed to consolidate past knowledge with generated pseudo samples. However, most existing generative replay methods use only a single task-specific token to control their models. This scheme is usually not strong enough to constrain the generative model due to insufficient information involved. In this paper, we propose a novel method, \textit{prompt conditioned VAE for lifelong learning} (PCLL), to enhance generative replay by incorporating tasks' statistics. 
PCLL captures task-specific distributions with a conditional variational autoencoder, conditioned on natural language prompts to guide the pseudo-sample generation. Moreover, it leverages a distillation process to further consolidate past knowledge by alleviating the noise in pseudo samples. Experiments on natural language understanding tasks of ToD systems demonstrate that PCLL significantly outperforms competitive baselines in building lifelong learning models.
We release the code and data at \href{https://github.com/AlibabaResearch/DAMO-ConvAI/tree/main/pcll}{GitHub}.
\end{abstract}

\section{Introduction}
Task-oriented dialogue (ToD) systems are of great importance in advanced AI applications \citep{zhang2020recent,dai2020learning,dai2021preview,he2022space,he2022unified,he2022galaxy}. However, most existing ToD systems are developed under the assumption that the data distribution remains unchanged \citep{pmtdst}. 
Unless the entire system is retrained, this setup may not be realistic when the ToD system deployed in practice needs to support new features and provides more services over time based on user demands. 
Without incurring the high cost of retraining, Lifelong Learning (LL) is able to acquire new knowledge continuously while preserving previously learned knowledge \citep{defy}. 
Hence, it's crucial to equip natural language understanding (NLU) modules, the vital components of ToD systems, with the lifelong learning ability.

The main issue for lifelong learning is \textit{catastrophic forgetting} \citep{forget2,parisi2019continual}, 
which refers to the phenomenon that a model forgets previously learned tasks when learning new tasks. 
Various approaches have been proposed to alleviate this issue \citep{ewc,mas,pnn,aljundi2017expert}. The replay-based methods are among the most effective and widely used ones \citep{icarl,deep_gr,dai2022lifelong}.
The main idea of replay-based methods is to
retrain samples or representations from already seen tasks when learning new tasks \citep{wholistic}. 
Some methods explicitly store previously seen real samples for replaying (\emph{experience replay}) \citep{icarl,chaudhry2019tiny}. However, this setting will be infeasible when data from previous tasks is unavailable due to data security concerns. 
Other methods try to generate pseudo samples using a generative model (\emph{generative replay}). This variant relieves the burden of storing previously seen data and has been widely adopted in previous studies \citep{defy,deep_gr,fearnet}.

The key to generative replay is to produce pseudo samples to approximate the real data distribution of previous tasks.
Intuitively, higher quality pseudo samples can better preserve learned tasks and lead to less forgetting in LL.
However, the generation of pseudo samples for each seen task in previous studies \citep{lamol,lamol-kd} is usually controlled by a single task-specific token.
It has been observed that this scheme is usually insufficient to constrain the PLM \citep{lamol}, due to limited information involved.
Consequently, the generated pseudo samples suffer from problems such as not being fluent or not corresponding well to the designated task.
Moreover, those special tokens are only introduced in the fine-tuning stage of the PLM.
This enlarges the gap between pre-training and fine-tuning of the PLM \citep{ppt} and harms the quality of the generated pseudo samples.
In addition, generated noisy pseudo samples may degenerate the LL performance.

To address the above issues, we propose a novel method, \underline{P}rompt \underline{C}onditioned VAE for \underline{L}ifelong \underline{L}earning (PCLL), to enhance generative replay on NLU tasks of ToD systems.
To impose strong control over the pseudo-sample generation, PCLL explicitly models latent task-specific distributions using a conditional variational autoencoder (CVAE) \citep{kingma2014auto,zhao2017learning}. Then it incorporates the corresponding task statistics to guide the generation of pseudo samples.
To reduce the gap between pretraining and finetuning, we construct natural language prompts to unify different NLU tasks while being specific to each task. These prompts not only contain meaningful semantics compared to special tokens, but also serve as conditions to assist CVAE in capturing task distributions.
Moreover, PCLL employs a knowledge distillation scheme to alleviate the impact of noisy pseudo samples during the replay process.
Leveraging the above strategies, PCLL can generate high-quality pseudo samples that better approximate the real distributions of previous tasks while tackling the aforementioned issues.

We validate our method on NLU tasks of ToD systems including both intent detection and slot filling. The results indicate that our approach generates high-quality pseudo samples and significantly outperforms competitive baselines.
Our main contributions are as follows,

\noindent \textbf{(1)} We propose a novel method, PCLL, to enhance generative replay for building lifelong NLU modules of ToD systems. 

\noindent \textbf{(2)} Conditioned on prompts, PCLL models latent task distributions with CVAE to guide the pseudo-sample generation and leverages knowledge distillation to further avoid forgetting.

\noindent \textbf{(3)} 
Our extensive experiments and comprehensive analyses demonstrate the superior performance of PCLL and the high quality of its generated samples.

\section{Related Work}
\subsection{Lifelong Learning}
There are generally three categories of LL methods:

\textbf{Regularization-based Methods}
aim to strike a balance between protecting already learned tasks while granting sufficient flexibility for a new task \citep{wholistic}.
Some methods \citep{ewc,mas,zenke2017continual,ebrahimi2019uncertainty} impose constraints on the modification of important weights.
Other methods introduce a distillation loss to constrain predicted features of the LL model. \citep{lwf,dhar2019learning,rannen2017encoder}.
However, these additional regularization terms may downgrade the model performance \citep{parisi2019continual}.

\textbf{Architecture-based Methods}
dedicate model parameters for each task to prevent forgetting \citep{defy}.
Some studies \citep{pathnet,HAT,hu2018overcoming} use static architectures and rely on task speciﬁc information to route through the architecture \citep{wholistic},
while other studies \citep{pnn,aljundi2017expert,zhai2020piggyback,tod,CTR,pruning,zhao2022semi} dynamically grow the architecture in the LL training process.
However, these methods either require capacity allocation for tasks at the beginning or are not feasible when model expansion is prohibited with limited resources \citep{lamol}.

\textbf{Replay-based Methods}
aim to preserve previous knowledge by replaying data from learned tasks.
One line of studies \citep{icarl,chaudhry2019tiny,gem,mifei,han2020continual,dst} keeps a small number of real samples from old tasks for replaying. However, these methods are unpractical when data from old tasks are unavailable.
Another line of studies \citep{deep_gr,fearnet,xiang2019incremental} utilizes a generative model to reproduce pseudo samples or representations from old tasks.

In this paper, we focus on improving generative replay, as it does not require allocating extra parameters or model capacity and can be used with any LL model.
Specifically, \citet{lamol} propose a general framework LAMOL for lifelong language learning to replay pseudo samples of previous tasks.
\citet{lamol-kd} improve LAMOL by training an extra teacher model before learning each new task, however, this increases the burden of the LL process.
\citet{rational-lamol} freeze critical parameters in LAMOL based on rationales, but those rationales are not always available for NLP tasks. 
All these previous works do not take task statistics into consideration, whereas our PCLL method incorporates the information of tasks' distributions to enhance generative replay.

\subsection{Prompt-based Learning in NLP}
Prompt-based learning has been found to be more effective than typical finetuning to use PLM \citep{schick-schutze-2021-just}. With prompts, we can convert various downstream tasks to a unified language modeling task \citep{brown2020language,schick-schutze-2021-just}.
Prompts can be either manually designed \citep{petroni-etal-2019-language,bowen} or generated automatically \citep{shin-etal-2020-autoprompt,Jiang,gao-etal-2021-making}.
Some recent studies employ prompt tuning on continual learning for dialogue state tracking \citep{pmtdst} and few-shot learning \citep{lfpt5}. 

\section{Methodology}
\subsection{Problem Definition}
We aim to build an LL model to learn a stream of NLU tasks sequentially $\mathcal{T}^{T}=\{t\}_{t=1}^{T}$ in dialogue systems, where $T$ can be infinite potentially. For each task $t$, a set of samples $\mathcal{D}_{t} = \{(x_k,y_k)\}_{k=1}^{N_{t}}$ are drawn from its underlying data distribution. Here, $x_k$ denotes the input utterance, and $y_k$ denotes the output label of NLU. In intent detection tasks, $y_k$ is the intent label of $x_k$; 
in slot filling tasks, $y_k$ is the slot-value pairs contained in $x_k$. 
Our objective is to learn a model that can perform well on all seen tasks and forget as little as possible.

\subsection{Overview}

\begin{figure}[t]
    \centering
    \includegraphics[width=180px]{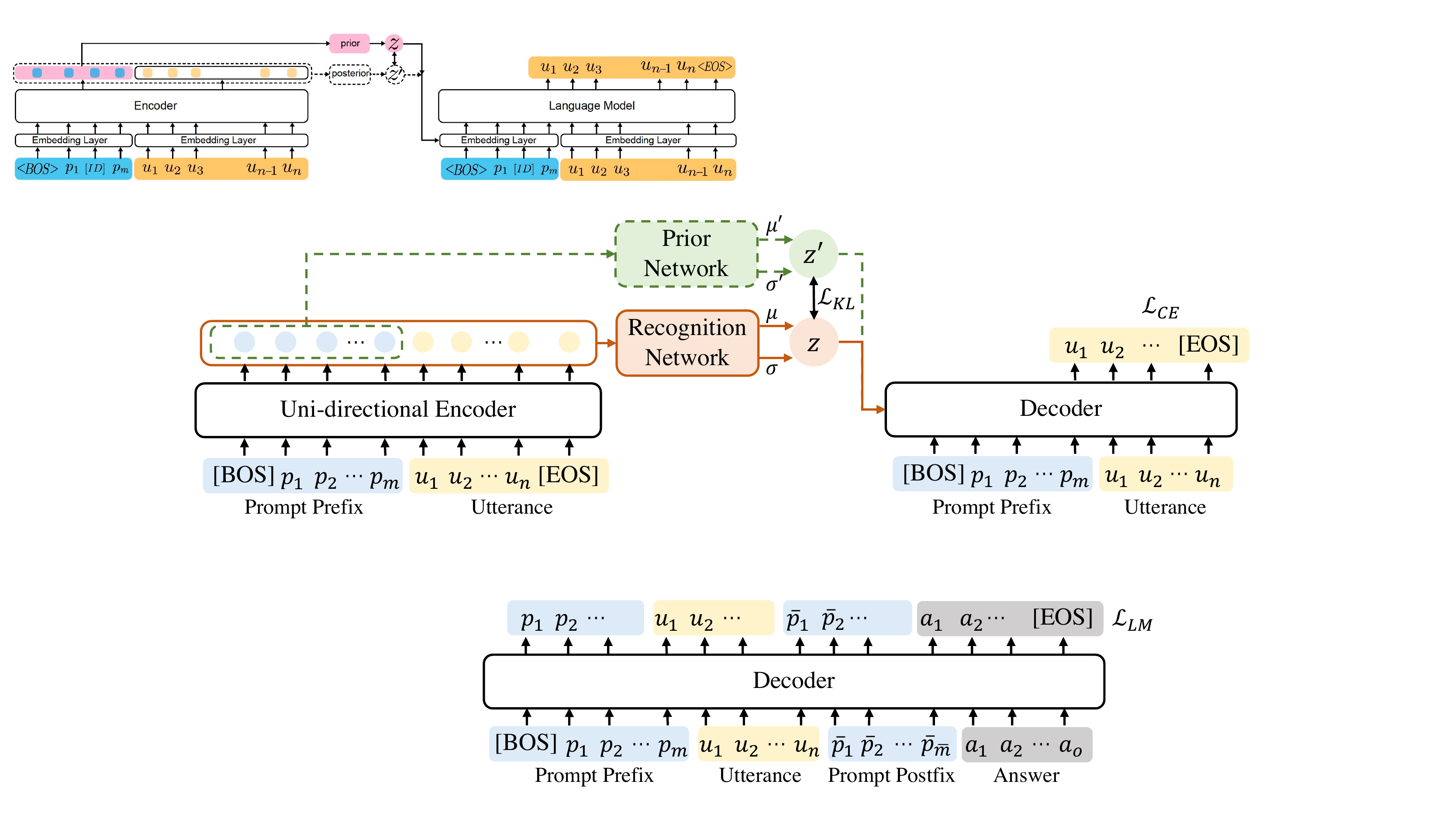}
    \caption{The training process of our model PCLL.}
    \label{fig:train_process}
\end{figure}

\begin{figure*}[htbp]
    \centering
    \includegraphics[width=420px]{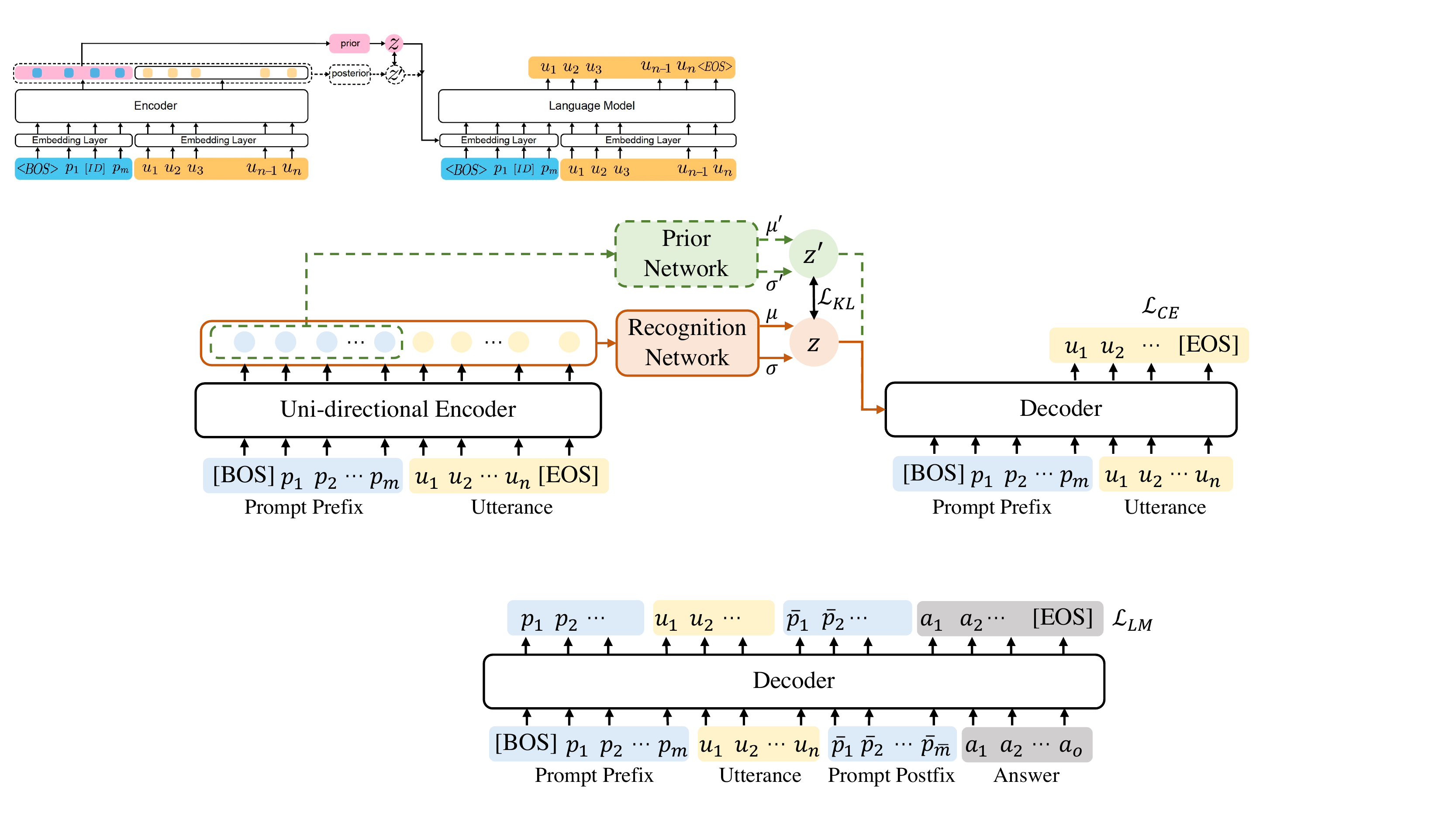}
    \caption{The architecture of the prompt conditioned VAE generator in PCLL. It captures the task distribution conditioned on prompts and incorporates the latent variable $z$ (or $z'$) into tokens' embeddings to guide the decoding.}
    \label{fig:cvae}
\end{figure*}

We start with a brief overview of our proposed PCLL method for generative replay (See Fig.~\ref{fig:train_process}).
PCLL consists of two components: an LM-based task solver to solve NLU tasks (Fig.~\ref{fig:lm}) and a CVAE-based generator (Fig.~\ref{fig:cvae}) to generate pseudo samples with the help of task-specific latent distributions. 
For the first task, PCLL is initialized with PLMs along with other parameters randomly initialized.
Before learning a new task $t$, we first use the PCLL model trained on previous tasks to generate pseudo samples for each of the learned tasks $\mathcal{T}^{t-1}$.
Then we interleave these pseudo samples with the training data in $\mathcal{D}_{t}$ and continue to train PCLL.
In this way, the model can learn the new task $t$ while consolidating the knowledge of past tasks. 

In the following sections, we first illustrate how PCLL learns the current task (Sec.~\ref{sec:prompts_for_NLU}, \ref{sec:cvae}). Then we describe the pseudo-sample generation process (Sec.~\ref{sec:sample_gen}), and finally, we introduce a knowledge distillation process to further improve the LL performance (Sec.~\ref{sec:distill}).

\subsection{LM-based Task Solver}\label{sec:prompts_for_NLU}
Following recent studies \citep{lamol,lamol-kd}, PCLL unifies different NLU tasks into a language modeling (LM) task and implements a task solver based on a PLM.
Different from previous studies that introduce randomly initialized special tokens in the fine-tuning stage \citep{lamol}, we construct task-specific natural language prompts for the solver.
These prompts carry rich semantic information to alleviate the mismatch between fine-tuning and pre-training of PLM.

For each input-output pair $(x, y)$ from task $t$,
our task solver is a LM that takes a prompt $g_t(x)$ as an input and predicts $y$. 
Specifically, $g_t(x)$ is constructed as $g_t(x) = g_t^{pre}\oplus x\oplus g_t^{post}$,
where $g_t^{pre}$ and $g_t^{post}$ are prompt prefix and postfix designed for task $t$, respectively, and $\oplus$ means the concatenation of word tokens.
For instance, if the task $t$ is an intent detection task,
we design $g_t(x)$ as: ``\texttt{For an utterance from the ID task, x has the following intent }'',
where ``\texttt{ID}'' represents the task name of $t$. 
After serializing the output $y$ into a token sequence, 
we can obtain a natural language sentence by simply concatenating $g_t(x)$ with $y$. We list detailed examples in Appendix ~\ref{appendix:input}.
Then the PLM $f_{\theta_t}$ for the current task $t$ is optimized on the concatenated sentence 
\begin{align}\label{eq:prompt}
\small
g_t(x,y)=g_t^{pre}\oplus x\oplus g_t^{post}\oplus y,
\end{align}
by maximizing the following objective (see Fig. \ref{fig:lm}):
\begin{align*}\label{eq:lm_loss}
\begin{split}
\mathcal{L}_{LM} = \log p_{\theta}(g_t(x,y)) + \lambda \log p_{\theta}(y | g_t(x)),
\end{split}
\end{align*}
in which the first term learns to decode the constructed sentence given the start token [BOS],
and the second term learns to predict the output $y$ after reading the prompt $g_t(x)$.
$\lambda$ is a scalar used to balance these two terms.

\subsection{Prompt Conditioned VAE Generator}\label{sec:cvae}
To construct high-quality pseudo-samples, PCLL leverages a CVAE module to build a pseudo-sample generator so that it can incorporate tasks' statistics to guide the generation of pseudo samples.
The CVAE module captures task-specific latent distributions by taking utterances as the input, conditioned on prefix prompts, and reconstructing the input during training.

Specifically, given an input utterance $x$ in task $t$, we assume a random variable $z$ captures the latent distribution over $x$.
We define a conditional distribution as $p(x, z| t) = p(x|z, t) p(z| t)$,
where we approximate $p(z| t)$ and $p(x|z, t)$ using deep neural networks with parameters $\phi$ and $\theta$, respectively.
We refer to $p_\phi(z|t)$ as the \textit{prior network} and $p_\theta(x | z, t)$ as the \textit{decoder}.
To reconstruct $x$, a latent variable $z$ is first sampled from $p_\phi(z|t)$ and then $x$ is decoded through $p_\theta(x | z, t)$.

In this study, we assume the prior of $z$ to be a multivariate Gaussian distribution with a diagonal covariance matrix,
and introduce a \textit{recognition network} $q_\psi(z | x, t)$ to approximate the intractable true posterior $p(z | x, t)$.
The goal of CVAE is to maximize the conditional log-likelihood $\log p(x | t) = \int p(x|z, t) p(z| t) d z$.
Employing variational inference, we can get the following evidence lower bound (ELBO) \citep{zhao2017learning} to maximize:
\begin{equation}
\small
\begin{aligned}
& \mathcal{L}_{\text{CVAE}} = \underbrace{\mathbb{E}_{q_{\psi}(z | x, t)} \log p_\theta(x | z, t)}_{\mathcal{L}_{\text{REC}}} \\
& - \beta \underbrace{\mathrm{KL}\left(q_\psi(z | x, t) \| p_\phi(z | t)\right)}_{\mathcal{L}_{\text{KL}}} \leq \log p(x | t),
\end{aligned}\label{eq:elbo}
\end{equation}
where $\beta$ is a scalar to balance the reconstruction term $\mathcal{L}_{\text{REC}}$ and the Kullback–Leibler (KL) divergence term $\mathcal{L}_{\text{KL}}$ and is adjusted by a cyclic annealing schedule \citep{fu2019cyclical} to alleviate the vanishing latent variable issue \citep{bowman2015generating}.
\paragraph{CVAE Implementation.}
When implementing each network in Eq.\ref{eq:elbo},
we use the prompt prefix $g_t^{pre}$ to represent the task $t$ because $g_t^{pre}$ involves the task name that can exclusively identify $t$.
Fig.~\ref{fig:cvae} shows the overall architecture of our PCLL model,
in which we use an unidirectional transformer \citep{transformer} to encode the concatenated sentence $g_t^{pre}\oplus x$ into hidden representations.
Then an attention-average block \citep{transformer_cvae} is introduced to pool the hidden representations of $g_t^{pre}$ and $g_t^{pre}\oplus x$ to single vectors, which are
further fed into a prior network $p_\phi(z|t)$ and recognition network $q_\psi(z | x, t)$ respectively.
Next, the reparametrization trick \citep{kingma2014auto} is used to obtain latent variables $z$ from the prior and posterior distributions.
Then $z$ is injected to the decoder $p_\theta(x | z, t)$ by adding to each token embedding (word embedding and position embedding, elementwisely) of the prompt \citep{transformer_cvae,optimus}.

In PCLL, the decoder $p_\theta(x | z, t)$ shares the same parameters with the PLM-based task solver $f_\theta$. This allows us to inherit the advantage of PLM and leverage a unified model to solve each task and generate pseudo samples simultaneously.

\subsection{Pseudo Sample Generation}\label{sec:sample_gen}
Generating pseudo samples for learned tasks involves two steps:
(1) PCLL generates a pseudo input utterance $x$ guided by a latent task distribution using the CVAE-based generator.
Specifically, for each seen task $t',~(t'<t)$, the model samples a latent variable $z_{t'}$ from the prior network $p_\phi(z_{t'}|t')$ with the constructed prompt prefix $g_{t'}^{pre}$ as the input.
Then the decoder takes $z_{t'}$ and $g_{t'}^{pre}$, and decodes them into the pseudo input $x$ using top-k sampling\footnote{Using other diversity enhanced decoding scheme may help produce more diverse pseudo samples \cite{wang2021diversifying}. We leave it for future works.} \cite{holtzman2019curious}.
(2) PCLL generates the output $y$ associated with $x$ using the solver (i.e., following Fig. \ref{fig:lm}).

\begin{figure}[t]
    \centering
    \includegraphics[width=220px]{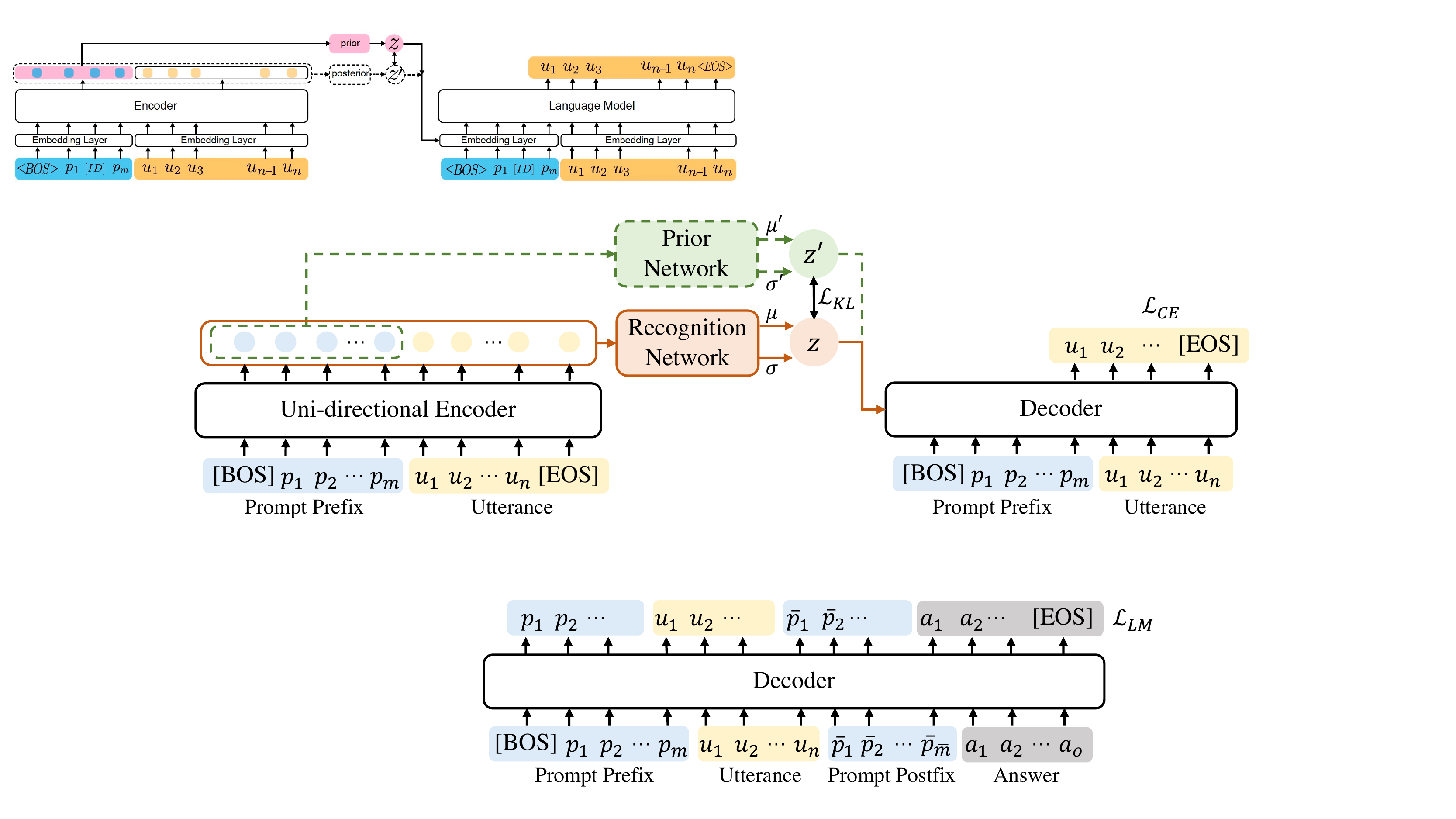}
    \caption{The LM-based solver for NLU tasks.
    The input-output pair $(x,y)$ is converted into a natural language prompts with $g_t^{pre}$ and $g_t^{post}$.}
    \label{fig:lm}
\end{figure}

\subsection{Knowledge Distillation}\label{sec:distill}
Previous generative replay approaches indistinguishably interleave pseudo data with the current task's training data.
However, this naive approach hurts the model performance since these pseudo data may contain noise and may drift from the real data distribution.
In this study, we utilize a knowledge distillation (KD) \citep{kd} process to prevent our model from being affected by these noisy pseudo data.

When training on a new task $t$,
we treat the model obtained on previous tasks $\mathcal{T}^{t-1}$ as a fixed teacher model $f_{\theta_\text{Tch}}$.
For each input-output pair $(x,y)$ in the pseudo data, 
$f_{\theta_{\text{Tch}}}$ is distilled on the generated pseudo data to the current model $f_\theta$ (i.e., serves as the student model) by maximizing the token-level distillation objective:
\begin{align*}
\begin{split}
 & \mathcal{L}_{\text{LM}}^{\text{KD}} =
 \scriptstyle\sum\limits_{l=1}^{|g_t(x,y)|}\sum\limits_{v \in \mathcal{V}} p_{\theta_\text{Tch}}(v | g_t(x,y)_{<l}) \log p_{\theta} (v | g_t(x,y)_{<l}) \\
 \!\!\!& + \scriptstyle \sum\limits_{l=1}^{|y|} \sum\limits_{v \in \mathcal{V}} p_{\theta_\text{Tch}}(v | g_t(x), y_{<l}) \log p_{\theta} (v | g_t(x), y_{<l}),\!\!
\end{split}
\end{align*}
where $g_t(x,y)<l$ and $y<l$ refers to the token sequence before the $l$-th token in $g_t(x,y)$ and $y$, respectively.
$\mathcal{V}$ represents the vocabulary set.

Similarly, when training the CVAE module,
we replace the reconstruction term $\mathcal{L}_{REC}$ of in Eq. \ref{eq:elbo} with a distillation objective:
\begin{align*}
\begin{split}
  \mathcal{L}_{\text{REC}}^{\text{KD}} = \mathop{\mathbb{E}}\limits_{q_{\psi}(z | x, t)} \scriptstyle \sum\limits_{l=1}^{|x|} \sum\limits_{v \in \mathcal{V}} & p_{\theta_\text{Tch}}(v |z,t, x_{<l}) \times \\ 
  & \log p_{\theta} (v |z, t, x_{<l}),
\end{split}
\end{align*}
and thus we maximize the following objective over the pseudo data $\mathcal{L}_{\text{CVAE}}^{\text{KD}} = \mathcal{L}_{\text{REC}}^{\text{KD}} - \beta \mathcal{L}_{\text{KL}}$.

Using the above KD strategy, the distributions produced by the teacher model contain richer knowledge compared to one-hot labels \citep{kd}.
These distributions constrain the student model (i.e., $f_\theta$) by preventing its weights from drifting too far when learning new tasks, thereby mitigating forgetting in lifelong learning.

Fig.\ref{fig:train_process} illustrates the training process of PCLL.
Specifically, when learning a new task $t$,
we optimize PCLL on training samples of $t$ with the following objective:
$\mathcal{L}_{\text{\text{LM}}} + \mathcal{L}_{\text{CVAE}}$.
For pseudo samples of previous tasks $t', (t'<t)$, we optimize the loss 
\begin{equation*}
\mathcal{L} = \alpha (\mathcal{L}_{\text{LM}}^{\text{KD}} + \mathcal{L}_{\text{CVAE}}^{\text{KD}}) + (1-\alpha)(\mathcal{L}_{\text{LM}}+\mathcal{L}_{\text{CVAE}}),
\end{equation*}
where $\alpha \in [0,1]$ is a scalar used to adjust knowledge distillation terms.


\section{Experiments}
\subsection{Datasets}
We evaluate the PCLL method on intent detection and slot filling based on public NLU benchmarks:

For intent detection, we collect six datasets that carry intent annotations:
HWU \citep{hwu}, BANKING \citep{banking}, CLINC \citep{clinc}, SNIPS \citep{snips}, AITS \citep{atis}, and TOP \citep{top}.
The dataset TOP is divided into three disjoint subsets TOP-S1, TOP-S2, and TOP-S3,
and these three subsets along with the other five datasets are regarded as separate LL tasks to increase the total number of tasks for sequential training.
Finally, we have eight tasks to be learned sequentially for this intent detection experiment.

For slot filling, we adopt five datasets that provide slot labels:
SNIPS, AITS, DSTC \citep{dstc}, MIT-MOVIE, and MIT-RESTAURANT \footnote{\url{groups.csail.mit.edu/sls/downloads}}.
Each dataset above is regarded as a separate LL task,
and thus five tasks are learned in lifelong slot filling experiments. More descriptions about datasets are in Appendix \ref{appendix:dataset}.

\subsection{Implementation Details}
We use the pretrained 12-layer GPT2 model \citep{gpt2} to initialize the encoder and decoder of our CVAE model.
The prior network and the recognition network are both set to be a 2-layer MLP with hidden size of 128.
When learning a new task $t$, PCLL balances the training data of $t$ and pseudo samples by generating 
$\gamma N_{t}$ pseudo samples for previously learned tasks.
$\gamma$ is the sampling ratio and $\gamma$ is set to 0.2 in our experiment following \citet{lamol}.
Each task for intent detection and slot filling is trained for 5 and 10 epochs, respectively.
We train PCLL on six random permutations of the task order. See Appendix \ref{appendix:six_orders} and \ref{appen:details} for more details.

\subsection{Baselines}
We compare PCLL with the following baselines:
\textbf{Fine-tune} directly fine-tunes the model on the task stream without preventing catastrophic forgetting;
\textbf{EWC} \citep{ewc} and \textbf{MAS} \citep{mas} are two regularization methods that mitigate forgetting by penalizing changes of important parameters for learned tasks;
\textbf{LAMOL-g} and \textbf{LAMOL-t} \citep{lamol} are two variants of the generative replay meth\textbf{}od LAMOL that control the generation of pseudo samples either using a global special token (LAMOL-g) or task-specific special tokens (LAMOL-t);
\textbf{L2KD} \cite{lamol-kd} improves LAMOL by assigning an extra teacher for each new task to perform knowledge distillation;
\textbf{ER} \citep{er} preserves previously seen real samples for replay to prevent forgetting.
We also consider some architecture-based baselines:
\textbf{HAT} \citep{HAT} creates a task-based hard attention during training;
\textbf{CTR} \citep{CTR} inserts continual learning plug-ins into BERT to mitigate forgetting and encourage knowledge transfer;
\textbf{Adapter} \citep{tod} builds residual adapter for each task independently. 
Since works in \citet{dst} and \citet{lfpt5} are specially designed for dialogue state tracking and few-shot learning, respectively, we do not consider them as our baselines.

Besides the above baselines, we further evaluate the model performance when all tasks are trained simultaneously in a multitask learning setting (\textbf{Multi}), which is often seen as an upper bound of LL.
For fair comparisons, all baselines are implemented following either the settings of \citet{lamol}, or their own reported settings.
For ER, we store 1\% of previously seen samples in memory following the setting of \citet{tod}.

\subsection{Evaluation Metrics}
We use the accuracy score, and macro-averaged F1 score \cite{coope2020span} to evaluate the performance of intent detection and slot filling tasks, respectively.
Moreover, we consider access to a test set for each of the $T$ tasks to learn in the LL process,
and define $R_{i,j}$ as the test score of the task $j$ after finishing learning the task $i$.
We follow previous studies \citet{gem,metric} to use the following two metrics to evaluate the performance of LL:
(1) \textbf{Average Score} (\textit{\textbf{Score}}) is defined as the average test score of all $T$ tasks after the LL process: $\mathrm{Score} = \frac{1}{T}\sum _{j=1}^T R_{T,j}$.
(2) \textbf{Learning Curve Area} (\textit{\textbf{LCA}}) is the area under the $Z_b$ curve, which captures the model's performance on all $T$ tasks \citep{chaudhry2018efficient}.
Specifically, $Z_b$ is the average score for all seen tasks at the training step $b$.
Here, high \textit{Score} and high \textit{LCA} are preferred for a good LL model.

\subsection{Main Results}

\begin{table}[t]
\scalebox{0.8}{
\begin{tabular}{lccccc}
\toprule
\multirow{2}{*}{Methods} & \multicolumn{2}{c}{Intent Detection} & \multicolumn{2}{c}{Slot Filling}\\       
\cmidrule{2-5}
                         &  Score       & LCA    &  Score   &  LCA       \\
\midrule
Finetune              &  14.09          & 28.76   & 15.38   & 19.55  \\
EWC                   &  14.16          & 28.34  & 15.67   & 19.51   \\
MAS                   &  14.15          & 28.61  & 15.59   & 19.37   \\
L2KD                  &  35.22 	        & 61.78  & 44.16   & 39.94   \\ 
LAMOL-g               &  50.30          & 60.67  & 45.12   & 38.03 \\
LAMOL-t               &  51.81          & 67.97  & 44.83   & 37.58   \\
ER                    &  78.19          & 71.36  & 44.95   & 39.32   \\
HAT                   &  73.92          & 73.03    & 61.99   & 67.33    \\
CTR                   &  67.44          & 71.11    & 63.84   & 67.28 \\
Adapter               &  81.15          & 75.60  & 58.21   & 48.47 \\
\midrule
PCLL                  & \textbf{90.25} & \textbf{88.82} & \textbf{74.48} & \textbf{68.41}       \\
\midrule
Multi (Upper Bound)   &  96.25 & N/A & 80.80 & N/A \\
\bottomrule
\end{tabular}}
\centering
\caption{Experiment results on both intent detection and slot filling tasks.
Each result is an average of six random task orders.
The best results among LL models are bold.
Our model PCLL is significantly better than other LL baselines with $p$-value $< 0.05$ under $t$-test.}
\label{tab:main_res}
\end{table} 

\begin{figure}[t]
  \centering
  \includegraphics[width=0.48\textwidth]{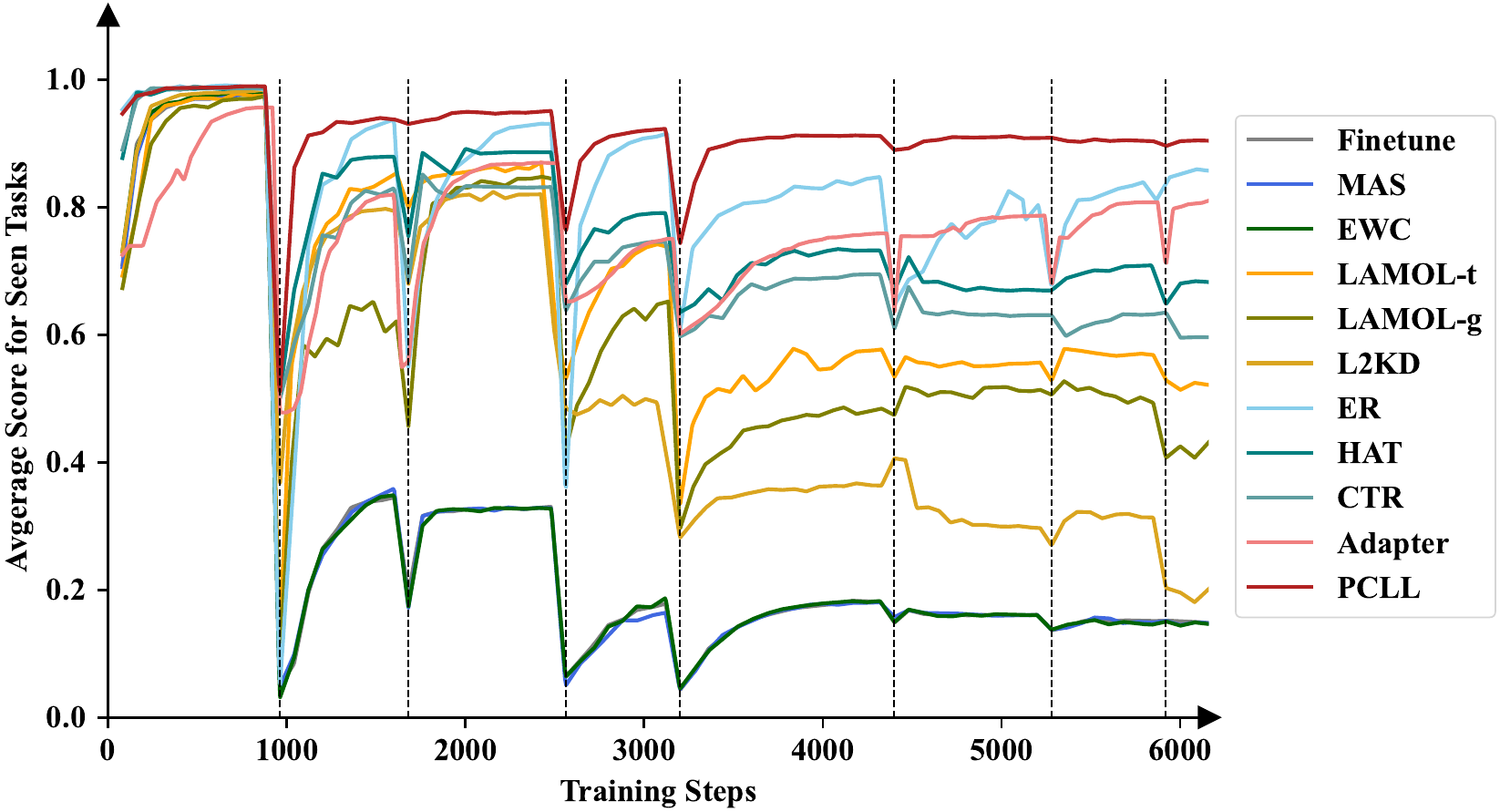}
  \caption{Learning curves of different methods on intent detection tasks. The dotted lines mean task switching.}
  \label{fig:avg_intent}
\end{figure}

Table \ref{tab:main_res} shows the performances of our model PCLL and all the baselines.
Our method PCLL significantly outperforms all baselines by a large margin on both intent detection and slot filling tasks.
To better understand the LL process,
we also plot the curve of the average score for all the models when trained using the same task order (see Fig. \ref{fig:avg_intent}).
From those results, we can observe that:

\noindent(1) Regularization-based methods (EWC and MAS) suffer from serious catastrophic forgetting, consistent with the observation of \citet{tod}.

\noindent(2)  Generative replay methods LAMOL-g, LAMOL-t, and L2KD alleviate the forgetting issue to some extent. 
However, replaying real samples (i.e., ER) performs much better.
This indicates that the quality of samples used for replaying is critical to addressing catastrophic forgetting, which matches our motivation to improve generative replay by generating high-quality pseudo samples.
Our method PCLL achieves higher performance than ER, indicating that PCLL can generate high-quality pseudo samples under the guidance of task distributions. Our analyses in Sec.~\ref{sec:qual_pseudo_sample} further prove this claim.

\noindent(3) Architecture-based methods HAT, CTR, and Adapter achieve good performance. However, PCLL still outperforms these baselines. This further validates the effectiveness of PCLL. Note that replay-based methods such as PCLL can be used together with these architecture-based methods to further improve the LL performance.

\noindent(4) From Fig~\ref{fig:avg_intent}, we can notice that when switching to new tasks, PCLL retains more knowledge about previous tasks (less performance degradation) compared to the baselines. This suggests that PCLL has a better ability to consolidate knowledge and mitigate catastrophic forgetting for LL.
\subsection{Ablation Studies}
We conduct ablation studies to verify the effectiveness of each proposed component in PCLL.
(1) \textbf{w/o Latent} means no latent distribution is modeled for each task,
i.e., the CVAE model in Section \ref{sec:cvae} is removed,
and pseudo samples are generated by directly feeding the prompt prefix into the LM $f_\theta$ without incorporating task-specific statistics.
(2) \textbf{w/o Task ID} means no task indicators are involved in the prompts. 
In other words, we design a task-independent prompt prefix by replacing the task ID with a general description ``current task'' (see Appendix \ref{appendix:input} for more details). In this way, the CVAE model degenerates to a VAE model that captures a global latent space for all tasks.
(3) \textbf{w/o KD} means that the knowledge distillation process in Section \ref{sec:distill} is not applied.
\begin{table}[b]
\small
\setlength\tabcolsep{2.2pt} 
\renewcommand\arraystretch{1.1}
\centering
\begin{tabular}{lcccc}
\toprule
             & \multicolumn{2}{c}{Intent Detection} & \multicolumn{2}{c}{Slot Filling}  \\ 
             \cline{2-3}  \cline{4-5} 
             & Score    & LCA & Score & LCA \\ 
\midrule
PCLL         & \textbf{90.25}  & \textbf{88.82} & \textbf{74.48}   & \textbf{68.41}     \\  
\midrule    
w/o Latent   & {86.09}  & 54.59  & 74.11     & 66.62       \\ 
w/o Task ID  & {72.37}  & 87.17  & 66.40     & 65.76     \\ 
w/o KD       & {81.63}  & 87.46  & 32.90     & 47.91 \\ 
\bottomrule
\end{tabular}
\caption{Ablation studies on two NLU tasks. Each result is an average of 6 random task orders.}
\label{tab:ablation}
\end{table}

From Table ~\ref{tab:ablation}, we can see that:
(1) Capturing task-specific latent distributions and incorporating them in the pseudo-sample generation process is crucial for building better LL models (\textbf{w/o Latent}).
(2) Using task-specific prompts helps to generate high-quality pseudo samples, thereby improving the LL performance (\textbf{w/o Task ID}).
(3) The proposed knowledge distillation process does mitigate the effects of noisy pseudo-samples and is beneficial for consolidating previously learned knowledge to prevent forgetting (\textbf{w/o KD}).

\section{Analyses}

\subsection{Soft Prompts vs. Manual Prompts}
We conduct analyses on soft prompts by replacing manually designed prompts with soft tokens in PCLL. 
Specifically, the prompt prefix $g_t^{pre}$ and postfix $g_t^{post}$ in Eq. \ref{eq:prompt} are replaced by several randomly initialized task-specific soft (learnable) tokens \cite{liu2021pre}. We also vary the lengths of these soft prompts to analyze their behaviors.

Results in Table~\ref{tab:softpmts} show that:
(1) Longer prefix prompts (i.e.\ more parameters guiding the pseudo-sample generation) generally lead to better LL performance; 
(2) Longer postfix prompts may not always lead to better LL performance. This may be because the postfix prompts are less important than prefix prompts since they do not participate in the pseudo-sample generation. Longer postfix prompts may bring in more noise, degenerating the performance;
(3) Using manual prompts in PCLL outperforms all its soft-prompt variants even though some soft prompts are much longer than manual prompts.
This justifies our claim that manual prompts carrying rich semantic information help to alleviate the mismatch between fine-tuning and pre-training of PLM and capture tasks' distributions,
and thus mitigate catastrophic forgetting in lifelong learning.

\begin{table}[t]
\centering
\small
\begin{tabular}{cllcc}
\toprule
& \multicolumn{2}{c}{Different Lengths} & \multicolumn{1}{c}{Score} & LCA \\ 
\midrule
\multirow{5}{*}{PCLL-Soft}
&                              & \#prefix=1                        & 51.30 & 57.77 \\
&                              & \#prefix=25                       & 83.41 & 75.99 \\
& \multirow{-3}{*}{\#postfix=1}   & \#prefix=100                      & 89.47 & 82.66 \\
\cmidrule{2-5}
&                              & {\#postfix=1} & 83.41 & 75.99 \\
&                              & {\#postfix=25} & 74.04 & 75.29 \\
& \multirow{-3}{*}{\#prefix=25} & {\#postfix=50} & 79.76 & 79.79 \\
\midrule
PCLL & \#prefix=15 & \#postfix=1  &  \textbf{90.25} & \textbf{88.82} \\
\bottomrule
\end{tabular}
\caption{Applying soft prompts on lifelong intent detection tasks. \#prefix and \#postfix indicate the lengths of prefix and postfix prompts, respectively. Each result is an average of 6 random task orders.}
\label{tab:softpmts}
\end{table}

\subsection{Manual Prompts} 
\paragraph{Different Designs.} We validate different designs of manual prompts in PCLL.
Specifically, we implement five different prompt templates with different lengths (Appendix \ref{appen:difpmt}). 
We observe that different manual prompts yield almost the same performance. This indicates that our method is robust to the design of manual prompts. (See Table \ref{tab:difpmts} in the Appendix).

\paragraph{Visualization of Attentions.} We provide the visualization of the attention scores over several manual prompts employed by PCLL. High attention scores of task names in Fig.~\ref{fig:attn_case} indicate that the task indicators play an important role in our manually designed prompts (see Appendix~\ref{appen:attn}).
\subsection{Qualities of Pseudo Samples}\label{sec:qual_pseudo_sample}
We validate the quality of pseudo samples generated by PCLL and all our generative replay baselines on intent detection tasks.
We use the distinct score \textbf{Dist-n} \citep{dist} to measure the proportion of unique n-grams in the generated pseudo samples' inputs ($n$=1,2,3,4).
Higher Dist-n indicates more diverse generated pseudo samples,
which is usually preferred because diverse samples help to approximate task distributions.
As shown in Table \ref{tab:dist}, PCLL can generate more diverse pseudo samples compared to other generative replay methods.
This demonstrates that pseudo samples constructed by our method are closer to real samples.

Further, we measure whether the generated pseudo samples can restore the distribution of real samples by visualizing samples' feature space with t-SNE \citep{tsne}.
As shown in Fig.~\ref{fig:tsne}, pseudo samples generated by PCLL are clustered in a similar pattern compared to real samples,
while those of LAMOL-t are scattered in the feature space. It shows that the pseudo samples generated by PCLL share closer distribution with the real samples compared to our baselines (see Appendix ~\ref{appen:pseudo-quality} for more details).
\begin{table}[t]
\centering
\scalebox{0.85}{
\begin{tabular}{lcccc}
\toprule
        & Dist-1 & Dist-2 & Dist-3  & Dist-4  \\ 
\midrule
LAMOL-g & 0.0602 & 0.2466 & 0.4489  & 0.6178 \\ 
LAMOL-t & 0.1758 & 0.4733 & 0.6837 	& 0.8090  \\ 
PCLL    & 0.2836 & 0.6566 & 0.8369 	& 0.9221 \\ 
\midrule
Real Sample & 0.4000 & 0.7972 & 0.9255 & 0.9717 \\
\bottomrule
\end{tabular}}
\caption{Distinct scores for generated pseudo samples.}
\label{tab:dist}
\end{table}

\subsection{Analyses of Latent Variables}
\begin{figure}
\small
    \centering
    \includegraphics[width=130px]{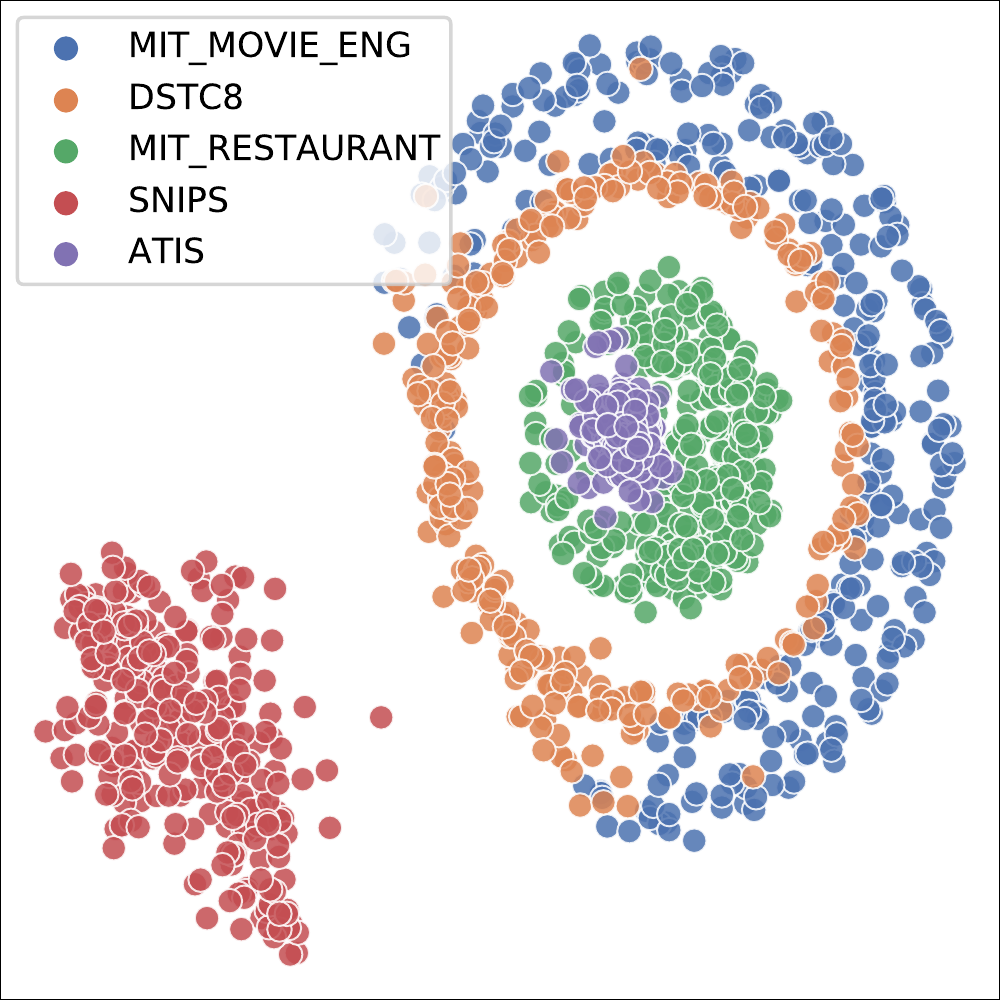}
    \caption{t-SNE visualization of latent variables.}
    \label{fig:latent_tsne}
\end{figure}

To further analyze the behavior of the pseudo sample generator, we visualize the latent space captured by the recognition network on slot filling tasks.
Specifically, for each sample in the test dataset, we extract a latent variable $z$ from its posterior distribution and use the t-SNE algorithm \citep{tsne} to visualize these variables in 2D space.
It can be seen from Figure \ref{fig:latent_tsne} that the latent spaces of different tasks are well clustered and clearly separated. This indicates that the latent variable $z$ is able to capture task-specific knowledge among learned tasks. 

We also analyze the influence of dimensions for latent variable $z$. The results are listed in Table ~\ref{tab:appen_difz}. We can notice that when we select the dimension of $z$ as 128, it can reach the best performance. 
This phenomenon is reasonable, when the dimension of $z$ is small, it may not catch enough information to model the task distribution; when the dimension is large, it may contain some noisy information, leading to poorer performance.
\begin{table}[t]
    \centering
    \begin{tabular}{ccc}
    \toprule
     $z$ dimension & \multicolumn{1}{c}{Score} & LCA \\ 
     \midrule
      32   & 90.00 & 88.27 \\
      128  & 90.25 & 88.82 \\
      256  & 90.10 & 88.30 \\
      512  & 90.04 & 88.26 \\
    \bottomrule
    \end{tabular}
    \caption{Analysis of different dimensions of the latent variable $z$ of PCLL on lifelong intent detection tasks. Each result is an average of six random task orders.}
    \label{tab:appen_difz}
\end{table}

\subsection{Influence of Sampling Ratio $\gamma$}
We analyze the influence of the sampling ratio $\gamma$ (ranging from 0.01 to 1.0) on the performance of PCLL. The results in Table~\ref{tab:ratio} indicate that PCLL is more effective in improving the LL performance when considering a small number of pseudo samples (See more details in Appendix~\ref{appen:sample_ratio}).

\subsection{Case Study}
We present several pseudo samples generated by PCLL and LAMOL in Table~\ref{tab:case} on the BANKING task for intent detection (see more cases in Appendix~\ref{appen:case}). 
We can observe that:
(1) Compared to LAMOL, pseudo samples produced by PCLL are closer to real samples from the BANKING dataset;
(2) Some samples generated by LAMOL are inconsistent with the task: LAMOL generates samples for the weather domain, which is not related to the BANKING task;
(3) LAMOL may also generate unmatched inputs and outputs in pseudo samples (last line in Table~\ref{tab:case}).
These observations verify our claim that a single task-specific token is too weak to constrain the PLM, and our method PCLL helps to generate high-quality pseudo samples that are consistent with each task.
\begin{table}[H]
\small
\setlength\tabcolsep{1.2pt} 
\centering
    \begin{tabular}{L{38pt}L{120pt}L{52pt}}
\toprule
 & \multicolumn{1}{c}{Input x} & \multicolumn{1}{c}{Output y} \\
 \midrule
\multirow{3}{*}{Real} & What exchange rate is it? & exchange rate\\
& My card never arrived. & card arrival \\
& I would like to reactivate my card. & card linking \\
\midrule
\multirow{3}{*}{PCLL} & What is my exchange rate? &  exchange rate \\
& My card hasn't come in yet. & card arrival \\
& How do I activate my card? & card linking \\
\midrule
\multirow{3}{*}{LAMOL} & the weather forecast & GetWeather \\
& is it going to be on my card &  card arrival \\
& I bought a used car & card linking \\
\bottomrule
    \end{tabular}
\caption{Real samples and generated pseudo samples for the BANKING task.}
\label{tab:case}
\end{table}

\subsection{Analyses of Forgetting for PCLL}
We provide more fine-grained analyses for the forgetting 
issue based on findings when learning with our proposed method PCLL.
In Appendix~\ref{appen:forgetting}, we carry out the analyses from the following four aspects: (1) unbalanced classes in some tasks, (2) conflicted label spaces for different tasks, (3) noisy pseudo labels for generated samples and (4) the diversity of pseudo samples created by PCLL.

\section{Conclusion}
In this paper, we propose PCLL to enhance generative replay for addressing catastrophic forgetting of lifelong learning in building NLU modules of ToD systems.
To construct high-quality pseudo samples, PCLL captures task-specific distributions with a prompt conditioned VAE to guide the generation of pseudo samples. 
Empirical results on two NLU tasks and extensive analyses demonstrate the superior performance of PCLL and the high quality of its generated pseudo samples.
Currently, we do not consider lifelong learning in the low-resource setting where only limited labeled data are available.
In the future, we will extend our framework to lifelong few-shot learning.

\section*{Limitations}
Here are some limitations of our work:
\begin{itemize}[leftmargin=*]
\setlength{\itemsep}{0pt}
\setlength{\parsep}{0pt}
\item We have not investigated lifelong learning in the low-resource setting where only limited labeled data are available. 
In future works, we will consider combining PCLL with meta-learning \cite{zhao-etal-2022-improving} to extend our framework to a lifelong few-shot learning setting.
We will also extend previous studies by using unlabeled data \cite{zhang2020dialogue,zhao2022semi} to build lifelong learning dialogue models.
\item We have not considered architecture-based methods for lifelong learning. However, our method PCLL can be readily combined with the architecture-based approach by leveraging parameter-efficient modules (e.g., Adapter \citep{adapter,zhang2021unsupervised}, LoRA \citep{lora}) into the model architecture to further mitigate the catastrophic forgetting issue. We will explore this direction in the future.
\end{itemize}

\section*{Ethical Considerations}
All our experiments are conducted on public available datasets.
All metrics used in our paper are automatic and do not need manual labor.
There are no direct ethical concerns in our study.

\section*{Acknowledgement}
Research on this paper was supported by Alibaba Group through Alibaba Research Intern Program and Hong Kong Research Grants Council (Grant No. 16204920).

\bibliography{anthology,custom}

\begin{thebibliography}{81}
\expandafter\ifx\csname natexlab\endcsname\relax\def\natexlab#1{#1}\fi

\bibitem[{Aljundi et~al.(2018)Aljundi, Babiloni, Elhoseiny, Rohrbach, and
  Tuytelaars}]{mas}
Rahaf Aljundi, Francesca Babiloni, Mohamed Elhoseiny, Marcus Rohrbach, and
  Tinne Tuytelaars. 2018.
\newblock Memory aware synapses: Learning what (not) to forget.
\newblock In \emph{Proceedings of the European Conference on Computer Vision
  (ECCV)}, pages 139--154.

\bibitem[{Aljundi et~al.(2017)Aljundi, Chakravarty, and
  Tuytelaars}]{aljundi2017expert}
Rahaf Aljundi, Punarjay Chakravarty, and Tinne Tuytelaars. 2017.
\newblock Expert gate: Lifelong learning with a network of experts.
\newblock In \emph{Proceedings of the IEEE Conference on Computer Vision and
  Pattern Recognition}, pages 3366--3375.

\bibitem[{Bowman et~al.(2016)Bowman, Vilnis, Vinyals, Dai, Jozefowicz, and
  Bengio}]{bowman2015generating}
Samuel Bowman, Luke Vilnis, Oriol Vinyals, Andrew Dai, Rafal Jozefowicz, and
  Samy Bengio. 2016.
\newblock Generating sentences from a continuous space.
\newblock In \emph{Proceedings of The 20th SIGNLL Conference on Computational
  Natural Language Learning}, pages 10--21.

\bibitem[{Brown et~al.(2020)Brown, Mann, Ryder, Subbiah, Kaplan, Dhariwal,
  Neelakantan, Shyam, Sastry, Askell et~al.}]{brown2020language}
Tom Brown, Benjamin Mann, Nick Ryder, Melanie Subbiah, Jared~D Kaplan, Prafulla
  Dhariwal, Arvind Neelakantan, Pranav Shyam, Girish Sastry, Amanda Askell,
  et~al. 2020.
\newblock Language models are few-shot learners.
\newblock \emph{Advances in neural information processing systems},
  33:1877--1901.

\bibitem[{Casanueva et~al.(2020)Casanueva, Tem{\v{c}}inas, Gerz, Henderson, and
  Vuli{\'c}}]{banking}
I{\~n}igo Casanueva, Tadas Tem{\v{c}}inas, Daniela Gerz, Matthew Henderson, and
  Ivan Vuli{\'c}. 2020.
\newblock Efficient intent detection with dual sentence encoders.
\newblock \emph{arXiv e-prints}, pages arXiv--2003.

\bibitem[{Chaudhry et~al.(2018{\natexlab{a}})Chaudhry, Dokania, Ajanthan, and
  Torr}]{metric}
Arslan Chaudhry, Puneet~K Dokania, Thalaiyasingam Ajanthan, and Philip~HS Torr.
  2018{\natexlab{a}}.
\newblock Riemannian walk for incremental learning: Understanding forgetting
  and intransigence.
\newblock In \emph{Proceedings of the European Conference on Computer Vision
  (ECCV)}, pages 532--547.

\bibitem[{Chaudhry et~al.(2018{\natexlab{b}})Chaudhry, Ranzato, Rohrbach, and
  Elhoseiny}]{chaudhry2018efficient}
Arslan Chaudhry, Marc’Aurelio Ranzato, Marcus Rohrbach, and Mohamed
  Elhoseiny. 2018{\natexlab{b}}.
\newblock Efficient lifelong learning with a-gem.
\newblock In \emph{International Conference on Learning Representations}.

\bibitem[{Chaudhry et~al.(2019)Chaudhry, Rohrbach, Elhoseiny, Ajanthan,
  Dokania, Torr, and Ranzato}]{chaudhry2019tiny}
Arslan Chaudhry, Marcus Rohrbach, Mohamed Elhoseiny, Thalaiyasingam Ajanthan,
  Puneet~K Dokania, Philip~HS Torr, and Marc'Aurelio Ranzato. 2019.
\newblock On tiny episodic memories in continual learning.
\newblock \emph{arXiv preprint arXiv:1902.10486}.

\bibitem[{Chuang et~al.(2020)Chuang, Su, and Chen}]{lamol-kd}
Yung-Sung Chuang, Shang-Yu Su, and Yun-Nung Chen. 2020.
\newblock Lifelong language knowledge distillation.
\newblock In \emph{Proceedings of the 2020 Conference on Empirical Methods in
  Natural Language Processing (EMNLP)}, pages 2914--2924.

\bibitem[{Coope et~al.(2020)Coope, Farghly, Gerz, Vuli{\'c}, and
  Henderson}]{coope2020span}
Samuel Coope, Tyler Farghly, Daniela Gerz, Ivan Vuli{\'c}, and Matthew
  Henderson. 2020.
\newblock Span-convert: Few-shot span extraction for dialog with pretrained
  conversational representations.
\newblock In \emph{Proceedings of the 58th Annual Meeting of the Association
  for Computational Linguistics}, pages 107--121.

\bibitem[{Coucke et~al.(2018)Coucke, Saade, Ball, Bluche, Caulier, Leroy,
  Doumouro, Gisselbrecht, Caltagirone, Lavril et~al.}]{snips}
Alice Coucke, Alaa Saade, Adrien Ball, Th{\'e}odore Bluche, Alexandre Caulier,
  David Leroy, Cl{\'e}ment Doumouro, Thibault Gisselbrecht, Francesco
  Caltagirone, Thibaut Lavril, et~al. 2018.
\newblock Snips voice platform: an embedded spoken language understanding
  system for private-by-design voice interfaces.
\newblock \emph{arXiv preprint arXiv:1805.10190}.

\bibitem[{Dai et~al.(2022)Dai, Lang, Zheng, Huang, Si, and
  Li}]{dai2022lifelong}
Yi~Dai, Hao Lang, Yinhe Zheng, Fei Huang, Luo Si, and Yongbin Li. 2022.
\newblock Lifelong learning for question answering with hierarchical prompts.
\newblock \emph{arXiv preprint arXiv:2208.14602}.

\bibitem[{Dai et~al.(2021)Dai, Li, Li, Sun, Huang, Si, and
  Zhu}]{dai2021preview}
Yinpei Dai, Hangyu Li, Yongbin Li, Jian Sun, Fei Huang, Luo Si, and Xiaodan
  Zhu. 2021.
\newblock Preview, attend and review: Schema-aware curriculum learning for
  multi-domain dialogue state tracking.
\newblock In \emph{Proceedings of the 59th Annual Meeting of the Association
  for Computational Linguistics and the 11th International Joint Conference on
  Natural Language Processing (Volume 2: Short Papers)}, pages 879--885.

\bibitem[{Dai et~al.(2020)Dai, Li, Tang, Li, Sun, and Zhu}]{dai2020learning}
Yinpei Dai, Hangyu Li, Chengguang Tang, Yongbin Li, Jian Sun, and Xiaodan Zhu.
  2020.
\newblock Learning low-resource end-to-end goal-oriented dialog for fast and
  reliable system deployment.
\newblock In \emph{Proceedings of the 58th Annual Meeting of the Association
  for Computational Linguistics}, pages 609--618.

\bibitem[{Delange et~al.(2021)Delange, Aljundi, Masana, Parisot, Jia,
  Leonardis, Slabaugh, and Tuytelaars}]{defy}
Matthias Delange, Rahaf Aljundi, Marc Masana, Sarah Parisot, Xu~Jia, Ales
  Leonardis, Greg Slabaugh, and Tinne Tuytelaars. 2021.
\newblock A continual learning survey: Defying forgetting in classification
  tasks.
\newblock \emph{IEEE Transactions on Pattern Analysis and Machine
  Intelligence}.

\bibitem[{Devlin et~al.(2019)Devlin, Chang, Lee, and
  Toutanova}]{devlin-etal-2019-bert}
Jacob Devlin, Ming-Wei Chang, Kenton Lee, and Kristina Toutanova. 2019.
\newblock \href {https://doi.org/10.18653/v1/N19-1423} {{BERT}: Pre-training of
  deep bidirectional transformers for language understanding}.
\newblock In \emph{Proceedings of the 2019 Conference of the North {A}merican
  Chapter of the Association for Computational Linguistics: Human Language
  Technologies, Volume 1 (Long and Short Papers)}, pages 4171--4186,
  Minneapolis, Minnesota. Association for Computational Linguistics.

\bibitem[{Dhar et~al.(2019)Dhar, Singh, Peng, Wu, and
  Chellappa}]{dhar2019learning}
Prithviraj Dhar, Rajat~Vikram Singh, Kuan-Chuan Peng, Ziyan Wu, and Rama
  Chellappa. 2019.
\newblock Learning without memorizing.
\newblock In \emph{Proceedings of the IEEE/CVF Conference on Computer Vision
  and Pattern Recognition}, pages 5138--5146.

\bibitem[{Ebrahimi et~al.(2019)Ebrahimi, Elhoseiny, Darrell, and
  Rohrbach}]{ebrahimi2019uncertainty}
Sayna Ebrahimi, Mohamed Elhoseiny, Trevor Darrell, and Marcus Rohrbach. 2019.
\newblock Uncertainty-guided continual learning with bayesian neural networks.
\newblock In \emph{International Conference on Learning Representations}.

\bibitem[{Fang et~al.(2021)Fang, Zeng, Liu, Bo, Dong, and
  Chen}]{transformer_cvae}
Le~Fang, Tao Zeng, Chaochun Liu, Liefeng Bo, Wen Dong, and Changyou Chen. 2021.
\newblock Transformer-based conditional variational autoencoder for
  controllable story generation.
\newblock \emph{arXiv e-prints}, pages arXiv--2101.

\bibitem[{Fernando et~al.(2017)Fernando, Banarse, Blundell, Zwols, Ha, Rusu,
  Pritzel, and Wierstra}]{pathnet}
Chrisantha Fernando, Dylan Banarse, Charles Blundell, Yori Zwols, David Ha,
  Andrei~A Rusu, Alexander Pritzel, and Daan Wierstra. 2017.
\newblock Pathnet: Evolution channels gradient descent in super neural
  networks.
\newblock \emph{arXiv preprint arXiv:1701.08734}.

\bibitem[{Fu et~al.(2019)Fu, Li, Liu, Gao, Celikyilmaz, and
  Carin}]{fu2019cyclical}
Hao Fu, Chunyuan Li, Xiaodong Liu, Jianfeng Gao, Asli Celikyilmaz, and Lawrence
  Carin. 2019.
\newblock Cyclical annealing schedule: A simple approach to mitigating kl
  vanishing.
\newblock In \emph{Proceedings of the 2019 Conference of the North American
  Chapter of the Association for Computational Linguistics: Human Language
  Technologies, Volume 1 (Long and Short Papers)}, pages 240--250.

\bibitem[{Gao et~al.(2021)Gao, Fisch, and Chen}]{gao-etal-2021-making}
Tianyu Gao, Adam Fisch, and Danqi Chen. 2021.
\newblock \href {https://doi.org/10.18653/v1/2021.acl-long.295} {Making
  pre-trained language models better few-shot learners}.
\newblock In \emph{Proceedings of the 59th Annual Meeting of the Association
  for Computational Linguistics and the 11th International Joint Conference on
  Natural Language Processing (Volume 1: Long Papers)}, pages 3816--3830,
  Online. Association for Computational Linguistics.

\bibitem[{Geng et~al.(2021)Geng, Yuan, Xu, Shen, Xu, and Yang}]{pruning}
Binzong Geng, Fajie Yuan, Qiancheng Xu, Ying Shen, Ruifeng Xu, and Min Yang.
  2021.
\newblock Continual learning for task-oriented dialogue system with iterative
  network pruning, expanding and masking.
\newblock In \emph{Proceedings of the 59th Annual Meeting of the Association
  for Computational Linguistics and the 11th International Joint Conference on
  Natural Language Processing (Volume 2: Short Papers)}, pages 517--523.

\bibitem[{Gu et~al.(2022)Gu, Han, Liu, and Huang}]{ppt}
Yuxian Gu, Xu~Han, Zhiyuan Liu, and Minlie Huang. 2022.
\newblock Ppt: Pre-trained prompt tuning for few-shot learning.
\newblock In \emph{Proceedings of the 60th Annual Meeting of the Association
  for Computational Linguistics (Volume 1: Long Papers)}, pages 8410--8423.

\bibitem[{Gupta et~al.(2018)Gupta, Shah, Mohit, Kumar, and Lewis}]{top}
Sonal Gupta, Rushin Shah, Mrinal Mohit, Anuj Kumar, and Mike Lewis. 2018.
\newblock Semantic parsing for task oriented dialog using hierarchical
  representations.
\newblock In \emph{EMNLP}.

\bibitem[{Han et~al.(2020)Han, Dai, Gao, Lin, Liu, Li, Sun, and
  Zhou}]{han2020continual}
Xu~Han, Yi~Dai, Tianyu Gao, Yankai Lin, Zhiyuan Liu, Peng Li, Maosong Sun, and
  Jie Zhou. 2020.
\newblock Continual relation learning via episodic memory activation and
  reconsolidation.
\newblock In \emph{Proceedings of the 58th Annual Meeting of the Association
  for Computational Linguistics}, pages 6429--6440.

\bibitem[{He et~al.(2022{\natexlab{a}})He, Dai, Hui, Yang, Cao, Dong, Huang,
  Si, and Li}]{he2022space}
Wanwei He, Yinpei Dai, Binyuan Hui, Min Yang, Zheng Cao, Jianbo Dong, Fei
  Huang, Luo Si, and Yongbin Li. 2022{\natexlab{a}}.
\newblock Space-2: Tree-structured semi-supervised contrastive pre-training for
  task-oriented dialog understanding.
\newblock In \emph{Proceedings of the 29th International Conference on
  Computational Linguistics}, pages 553--569.

\bibitem[{He et~al.(2022{\natexlab{b}})He, Dai, Yang, Sun, Huang, Si, and
  Li}]{he2022unified}
Wanwei He, Yinpei Dai, Min Yang, Jian Sun, Fei Huang, Luo Si, and Yongbin Li.
  2022{\natexlab{b}}.
\newblock Unified dialog model pre-training for task-oriented dialog
  understanding and generation.
\newblock In \emph{Proceedings of the 45th International ACM SIGIR Conference
  on Research and Development in Information Retrieval}, pages 187--200.

\bibitem[{He et~al.(2022{\natexlab{c}})He, Dai, Zheng, Wu, Cao, Liu, Jiang,
  Yang, Huang, Si et~al.}]{he2022galaxy}
Wanwei He, Yinpei Dai, Yinhe Zheng, Yuchuan Wu, Zheng Cao, Dermot Liu, Peng
  Jiang, Min Yang, Fei Huang, Luo Si, et~al. 2022{\natexlab{c}}.
\newblock Galaxy: A generative pre-trained model for task-oriented dialog with
  semi-supervised learning and explicit policy injection.
\newblock In \emph{Proceedings of the AAAI Conference on Artificial
  Intelligence}, volume~36, pages 10749--10757.

\bibitem[{Hemphill et~al.(1990)Hemphill, Godfrey, and Doddington}]{atis}
Charles~T Hemphill, John~J Godfrey, and George~R Doddington. 1990.
\newblock The atis spoken language systems pilot corpus.
\newblock In \emph{Speech and Natural Language: Proceedings of a Workshop Held
  at Hidden Valley, Pennsylvania, June 24-27, 1990}.

\bibitem[{Hinton et~al.(2015)Hinton, Vinyals, and Dean}]{kd}
Geoffrey Hinton, Oriol Vinyals, and Jeff Dean. 2015.
\newblock Distilling the knowledge in a neural network.
\newblock \emph{stat}, 1050:9.

\bibitem[{Holtzman et~al.(2019)Holtzman, Buys, Du, Forbes, and
  Choi}]{holtzman2019curious}
Ari Holtzman, Jan Buys, Li~Du, Maxwell Forbes, and Yejin Choi. 2019.
\newblock The curious case of neural text degeneration.
\newblock In \emph{International Conference on Learning Representations}.

\bibitem[{Houlsby et~al.(2019)Houlsby, Giurgiu, Jastrzebski, Morrone,
  De~Laroussilhe, Gesmundo, Attariyan, and Gelly}]{adapter}
Neil Houlsby, Andrei Giurgiu, Stanislaw Jastrzebski, Bruna Morrone, Quentin
  De~Laroussilhe, Andrea Gesmundo, Mona Attariyan, and Sylvain Gelly. 2019.
\newblock Parameter-efficient transfer learning for nlp.
\newblock In \emph{International Conference on Machine Learning}, pages
  2790--2799. PMLR.

\bibitem[{Hu et~al.(2021)Hu, Wallis, Allen-Zhu, Li, Wang, Wang, Chen
  et~al.}]{lora}
Edward~J Hu, Phillip Wallis, Zeyuan Allen-Zhu, Yuanzhi Li, Shean Wang, Lu~Wang,
  Weizhu Chen, et~al. 2021.
\newblock Lora: Low-rank adaptation of large language models.
\newblock In \emph{International Conference on Learning Representations}.

\bibitem[{Hu et~al.(2018)Hu, Lin, Liu, Tao, Tao, Ma, Zhao, and
  Yan}]{hu2018overcoming}
Wenpeng Hu, Zhou Lin, Bing Liu, Chongyang Tao, Zhengwei Tao, Jinwen Ma, Dongyan
  Zhao, and Rui Yan. 2018.
\newblock Overcoming catastrophic forgetting for continual learning via model
  adaptation.
\newblock In \emph{International Conference on Learning Representations}.

\bibitem[{Jiang et~al.(2020)Jiang, Xu, Araki, and Neubig}]{Jiang}
Zhengbao Jiang, Frank~F. Xu, Jun Araki, and Graham Neubig. 2020.
\newblock \href {https://doi.org/10.1162/tacl_a_00324} {{How Can We Know What
  Language Models Know?}}
\newblock \emph{Transactions of the Association for Computational Linguistics}.

\bibitem[{Kanwatchara et~al.(2021)Kanwatchara, Horsuwan, Lertvittayakumjorn,
  Kijsirikul, and Vateekul}]{rational-lamol}
Kasidis Kanwatchara, Thanapapas Horsuwan, Piyawat Lertvittayakumjorn, Boonserm
  Kijsirikul, and Peerapon Vateekul. 2021.
\newblock Rational lamol: A rationale-based lifelong learning framework.
\newblock In \emph{Proceedings of the 59th Annual Meeting of the Association
  for Computational Linguistics and the 11th International Joint Conference on
  Natural Language Processing (Volume 1: Long Papers)}, pages 2942--2953.

\bibitem[{Ke et~al.(2021)Ke, Liu, Ma, Xu, and Shu}]{CTR}
Zixuan Ke, Bing Liu, Nianzu Ma, Hu~Xu, and Lei Shu. 2021.
\newblock Achieving forgetting prevention and knowledge transfer in continual
  learning.
\newblock In \emph{Advances in Neural Information Processing Systems}.

\bibitem[{Kemker and Kanan(2018)}]{fearnet}
Ronald Kemker and Christopher Kanan. 2018.
\newblock Fearnet: Brain-inspired model for incremental learning.
\newblock In \emph{International Conference on Learning Representations}.

\bibitem[{Kingma and Welling(2014)}]{kingma2014auto}
Diederik~P Kingma and Max Welling. 2014.
\newblock Auto-encoding variational bayes.
\newblock \emph{stat}, 1050:10.

\bibitem[{Larson et~al.(2019)Larson, Mahendran, Peper, Clarke, Lee, Hill,
  Kummerfeld, Leach, Laurenzano, Tang et~al.}]{clinc}
Stefan Larson, Anish Mahendran, Joseph~J Peper, Christopher Clarke, Andrew Lee,
  Parker Hill, Jonathan~K Kummerfeld, Kevin Leach, Michael~A Laurenzano,
  Lingjia Tang, et~al. 2019.
\newblock An evaluation dataset for intent classification and out-of-scope
  prediction.
\newblock In \emph{Proceedings of the 2019 Conference on Empirical Methods in
  Natural Language Processing and the 9th International Joint Conference on
  Natural Language Processing (EMNLP-IJCNLP)}, pages 1311--1316.

\bibitem[{Li et~al.(2020)Li, Gao, Li, Peng, Li, Zhang, and Gao}]{optimus}
Chunyuan Li, Xiang Gao, Yuan Li, Baolin Peng, Xiujun Li, Yizhe Zhang, and
  Jianfeng Gao. 2020.
\newblock Optimus: Organizing sentences via pre-trained modeling of a latent
  space.
\newblock In \emph{Proceedings of the 2020 Conference on Empirical Methods in
  Natural Language Processing (EMNLP)}, pages 4678--4699.

\bibitem[{Li et~al.(2016)Li, Galley, Brockett, Gao, and Dolan}]{dist}
Jiwei Li, Michel Galley, Chris Brockett, Jianfeng Gao, and Bill Dolan. 2016.
\newblock A diversity-promoting objective function for neural conversation
  models.
\newblock In \emph{HLT-NAACL}.

\bibitem[{Li and Hoiem(2017)}]{lwf}
Zhizhong Li and Derek Hoiem. 2017.
\newblock Learning without forgetting.
\newblock \emph{IEEE transactions on pattern analysis and machine
  intelligence}, 40(12):2935--2947.

\bibitem[{Liu et~al.(2021{\natexlab{a}})Liu, Yuan, Fu, Jiang, Hayashi, and
  Neubig}]{liu2021pre}
Pengfei Liu, Weizhe Yuan, Jinlan Fu, Zhengbao Jiang, Hiroaki Hayashi, and
  Graham Neubig. 2021{\natexlab{a}}.
\newblock Pre-train, prompt, and predict: A systematic survey of prompting
  methods in natural language processing.
\newblock \emph{arXiv preprint arXiv:2107.13586}.

\bibitem[{Liu et~al.(2021{\natexlab{b}})Liu, Cao, Liu, Chen, Cai, Yang, He,
  Liu, and Zhao}]{dst}
Qingbin Liu, Pengfei Cao, Cao Liu, Jiansong Chen, Xunliang Cai, Fan Yang,
  Shizhu He, Kang Liu, and Jun Zhao. 2021{\natexlab{b}}.
\newblock Domain-lifelong learning for dialogue state tracking via knowledge
  preservation networks.
\newblock In \emph{Proceedings of the 2021 Conference on Empirical Methods in
  Natural Language Processing}, pages 2301--2311.

\bibitem[{Liu et~al.(2019)Liu, Eshghi, Swietojanski, and Rieser}]{hwu}
Xingkun Liu, Arash Eshghi, Pawel Swietojanski, and Verena Rieser. 2019.
\newblock Benchmarking natural language understanding services for building
  conversational agents.

\bibitem[{Lopez-Paz and Ranzato(2017)}]{gem}
David Lopez-Paz and Marc'Aurelio Ranzato. 2017.
\newblock Gradient episodic memory for continual learning.
\newblock \emph{Advances in neural information processing systems},
  30:6467--6476.

\bibitem[{Madotto et~al.(2021)Madotto, Lin, Zhou, Moon, Crook, Liu, Yu, Cho,
  Fung, and Wang}]{tod}
Andrea Madotto, Zhaojiang Lin, Zhenpeng Zhou, Seungwhan Moon, Paul~A Crook,
  Bing Liu, Zhou Yu, Eunjoon Cho, Pascale Fung, and Zhiguang Wang. 2021.
\newblock Continual learning in task-oriented dialogue systems.
\newblock In \emph{Proceedings of the 2021 Conference on Empirical Methods in
  Natural Language Processing}, pages 7452--7467.

\bibitem[{McClelland et~al.(1995)McClelland, McNaughton, and
  O'Reilly}]{forget2}
James~L McClelland, Bruce~L McNaughton, and Randall~C O'Reilly. 1995.
\newblock Why there are complementary learning systems in the hippocampus and
  neocortex: insights from the successes and failures of connectionist models
  of learning and memory.
\newblock \emph{Psychological review}, 102(3):419.

\bibitem[{Mi et~al.(2020)Mi, Chen, Zhao, Huang, and Faltings}]{mifei}
Fei Mi, Liangwei Chen, Mengjie Zhao, Minlie Huang, and Boi Faltings. 2020.
\newblock Continual learning for natural language generation in task-oriented
  dialog systems.
\newblock In \emph{Proceedings of the 2020 Conference on Empirical Methods in
  Natural Language Processing: Findings}, pages 3461--3474.

\bibitem[{Mundt et~al.(2020)Mundt, Hong, Pliushch, and Ramesh}]{wholistic}
Martin Mundt, Yong~Won Hong, Iuliia Pliushch, and Visvanathan Ramesh. 2020.
\newblock A wholistic view of continual learning with deep neural networks:
  Forgotten lessons and the bridge to active and open world learning.
\newblock \emph{arXiv e-prints}, pages arXiv--2009.

\bibitem[{Parisi et~al.(2019)Parisi, Kemker, Part, Kanan, and
  Wermter}]{parisi2019continual}
German~I Parisi, Ronald Kemker, Jose~L Part, Christopher Kanan, and Stefan
  Wermter. 2019.
\newblock Continual lifelong learning with neural networks: A review.
\newblock \emph{Neural Networks}, 113:54--71.

\bibitem[{Petroni et~al.(2019)Petroni, Rockt{\"a}schel, Riedel, Lewis, Bakhtin,
  Wu, and Miller}]{petroni-etal-2019-language}
Fabio Petroni, Tim Rockt{\"a}schel, Sebastian Riedel, Patrick Lewis, Anton
  Bakhtin, Yuxiang Wu, and Alexander Miller. 2019.
\newblock \href {https://doi.org/10.18653/v1/D19-1250} {Language models as
  knowledge bases?}
\newblock In \emph{Proceedings of the 2019 Conference on Empirical Methods in
  Natural Language Processing and the 9th International Joint Conference on
  Natural Language Processing (EMNLP-IJCNLP)}, Hong Kong, China. Association
  for Computational Linguistics.

\bibitem[{Qin and Joty(2022)}]{lfpt5}
Chengwei Qin and Shafiq Joty. 2022.
\newblock Lfpt5: A unified framework for lifelong few-shot language learning
  based on prompt tuning of t5.
\newblock \emph{ICLR 2022}.

\bibitem[{Radford et~al.(2019)Radford, Wu, Child, Luan, Amodei, Sutskever
  et~al.}]{gpt2}
Alec Radford, Jeffrey Wu, Rewon Child, David Luan, Dario Amodei, Ilya
  Sutskever, et~al. 2019.
\newblock Language models are unsupervised multitask learners.
\newblock \emph{OpenAI blog}, 1(8):9.

\bibitem[{Rannen et~al.(2017)Rannen, Aljundi, Blaschko, and
  Tuytelaars}]{rannen2017encoder}
Amal Rannen, Rahaf Aljundi, Matthew~B Blaschko, and Tinne Tuytelaars. 2017.
\newblock Encoder based lifelong learning.
\newblock In \emph{Proceedings of the IEEE International Conference on Computer
  Vision}, pages 1320--1328.

\bibitem[{Rastogi et~al.(2020)Rastogi, Zang, Sunkara, Gupta, and
  Khaitan}]{dstc}
Abhinav Rastogi, Xiaoxue Zang, Srinivas Sunkara, Raghav Gupta, and Pranav
  Khaitan. 2020.
\newblock Towards scalable multi-domain conversational agents: The
  schema-guided dialogue dataset.
\newblock In \emph{Proceedings of the AAAI Conference on Artificial
  Intelligence}, volume~34, pages 8689--8696.

\bibitem[{Rebuffi et~al.(2017)Rebuffi, Kolesnikov, Sperl, and Lampert}]{icarl}
Sylvestre-Alvise Rebuffi, Alexander Kolesnikov, Georg Sperl, and Christoph~H
  Lampert. 2017.
\newblock icarl: Incremental classifier and representation learning.
\newblock In \emph{Proceedings of the IEEE conference on Computer Vision and
  Pattern Recognition}, pages 2001--2010.

\bibitem[{Rolnick et~al.(2019)Rolnick, Ahuja, Schwarz, Lillicrap, and
  Wayne}]{er}
David Rolnick, Arun Ahuja, Jonathan Schwarz, Timothy~P Lillicrap, and Greg
  Wayne. 2019.
\newblock Experience replay for continual learning.
\newblock In \emph{Proceedings of the 33rd International Conference on Neural
  Information Processing Systems}, pages 350--360.

\bibitem[{Rusu et~al.(2016)Rusu, Rabinowitz, Desjardins, Soyer, Kirkpatrick,
  Kavukcuoglu, Pascanu, and Hadsell}]{pnn}
Andrei~A Rusu, Neil~C Rabinowitz, Guillaume Desjardins, Hubert Soyer, James
  Kirkpatrick, Koray Kavukcuoglu, Razvan Pascanu, and Raia Hadsell. 2016.
\newblock Progressive neural networks.
\newblock \emph{arXiv preprint arXiv:1606.04671}.

\bibitem[{Schick and Sch{\"u}tze(2021)}]{schick-schutze-2021-just}
Timo Schick and Hinrich Sch{\"u}tze. 2021.
\newblock \href {https://doi.org/10.18653/v1/2021.naacl-main.185} {It{'}s not
  just size that matters: Small language models are also few-shot learners}.
\newblock In \emph{Proceedings of the 2021 Conference of the North American
  Chapter of the Association for Computational Linguistics: Human Language
  Technologies}. Association for Computational Linguistics.

\bibitem[{Schwarz et~al.(2018)Schwarz, Czarnecki, Luketina, Grabska-Barwinska,
  Teh, Pascanu, and Hadsell}]{ewc}
Jonathan Schwarz, Wojciech Czarnecki, Jelena Luketina, Agnieszka
  Grabska-Barwinska, Yee~Whye Teh, Razvan Pascanu, and Raia Hadsell. 2018.
\newblock Progress \& compress: A scalable framework for continual learning.
\newblock In \emph{International Conference on Machine Learning}, pages
  4528--4537. PMLR.

\bibitem[{Serr{\`a} et~al.(2018)Serr{\`a}, Suris, Miron, and Karatzoglou}]{HAT}
Joan Serr{\`a}, Didac Suris, Marius Miron, and Alexandros Karatzoglou. 2018.
\newblock Overcoming catastrophic forgetting with hard attention to the task.
\newblock In \emph{ICML}.

\bibitem[{Shin et~al.(2017)Shin, Lee, Kim, and Kim}]{deep_gr}
Hanul Shin, Jung~Kwon Lee, Jaehong Kim, and Jiwon Kim. 2017.
\newblock Continual learning with deep generative replay.
\newblock In \emph{NIPS}.

\bibitem[{Shin et~al.(2020)Shin, Razeghi, Logan~IV, Wallace, and
  Singh}]{shin-etal-2020-autoprompt}
Taylor Shin, Yasaman Razeghi, Robert~L. Logan~IV, Eric Wallace, and Sameer
  Singh. 2020.
\newblock \href {https://doi.org/10.18653/v1/2020.emnlp-main.346}
  {{A}uto{P}rompt: {E}liciting {K}nowledge from {L}anguage {M}odels with
  {A}utomatically {G}enerated {P}rompts}.
\newblock In \emph{Proceedings of the 2020 Conference on Empirical Methods in
  Natural Language Processing (EMNLP)}, pages 4222--4235, Online. Association
  for Computational Linguistics.

\bibitem[{Sun et~al.(2020)Sun, Ho, and Lee}]{lamol}
Fan-Keng Sun, Cheng-Hao Ho, and Hung-Yi Lee. 2020.
\newblock Lamol: Language modeling for lifelong language learning.
\newblock In \emph{International Conference on Learning Representations}.

\bibitem[{Van~der Maaten and Hinton(2008)}]{tsne}
Laurens Van~der Maaten and Geoffrey Hinton. 2008.
\newblock Visualizing data using t-sne.
\newblock \emph{Journal of machine learning research}, 9(11).

\bibitem[{Vaswani et~al.(2017)Vaswani, Shazeer, Parmar, Uszkoreit, Jones,
  Gomez, Kaiser, and Polosukhin}]{transformer}
Ashish Vaswani, Noam Shazeer, Niki Parmar, Jakob Uszkoreit, Llion Jones,
  Aidan~N Gomez, {\L}ukasz Kaiser, and Illia Polosukhin. 2017.
\newblock Attention is all you need.
\newblock In \emph{Advances in neural information processing systems}, pages
  5998--6008.

\bibitem[{Wang et~al.(2021)Wang, Zheng, Jiang, and
  Huang}]{wang2021diversifying}
Yida Wang, Yinhe Zheng, Yong Jiang, and Minlie Huang. 2021.
\newblock Diversifying dialog generation via adaptive label smoothing.
\newblock In \emph{Proceedings of the 59th Annual Meeting of the Association
  for Computational Linguistics and the 11th International Joint Conference on
  Natural Language Processing (Volume 1: Long Papers)}, pages 3507--3520.

\bibitem[{Xiang et~al.(2019)Xiang, Fu, Ji, and Huang}]{xiang2019incremental}
Ye~Xiang, Ying Fu, Pan Ji, and Hua Huang. 2019.
\newblock Incremental learning using conditional adversarial networks.
\newblock In \emph{Proceedings of the IEEE/CVF International Conference on
  Computer Vision}, pages 6619--6628.

\bibitem[{Yu et~al.(2019)Yu, Zhang, Su, Wang, Liu, Wang, and Li}]{bowen}
Bowen Yu, Zhenyu Zhang, Jianlin Su, Yubin Wang, Tingwen Liu, Bin Wang, and
  Sujian Li. 2019.
\newblock \href {http://arxiv.org/abs/1909.04273} {Joint extraction of entities
  and relations based on a novel decomposition strategy}.
\newblock \emph{CoRR}, abs/1909.04273.

\bibitem[{Zenke et~al.(2017)Zenke, Poole, and Ganguli}]{zenke2017continual}
Friedemann Zenke, Ben Poole, and Surya Ganguli. 2017.
\newblock Continual learning through synaptic intelligence.
\newblock In \emph{International Conference on Machine Learning}, pages
  3987--3995. PMLR.

\bibitem[{Zhai et~al.(2020)Zhai, Chen, He, Nawhal, Tung, and
  Mori}]{zhai2020piggyback}
Mengyao Zhai, Lei Chen, Jiawei He, Megha Nawhal, Frederick Tung, and Greg Mori.
  2020.
\newblock Piggyback gan: Efficient lifelong learning for image conditioned
  generation.
\newblock In \emph{European Conference on Computer Vision}, pages 397--413.
  Springer.

\bibitem[{Zhang et~al.(2021)Zhang, Zheng, Mao, and
  Huang}]{zhang2021unsupervised}
Rongsheng Zhang, Yinhe Zheng, Xiaoxi Mao, and Minlie Huang. 2021.
\newblock Unsupervised domain adaptation with adapter.
\newblock \emph{arXiv preprint arXiv:2111.00667}.

\bibitem[{Zhang et~al.(2020{\natexlab{a}})Zhang, Zheng, Shao, Mao, Xi, and
  Huang}]{zhang2020dialogue}
Rongsheng Zhang, Yinhe Zheng, Jianzhi Shao, Xiaoxi Mao, Yadong Xi, and Minlie
  Huang. 2020{\natexlab{a}}.
\newblock Dialogue distillation: Open-domain dialogue augmentation using
  unpaired data.
\newblock In \emph{Proceedings of the 2020 Conference on Empirical Methods in
  Natural Language Processing (EMNLP)}, pages 3449--3460.

\bibitem[{Zhang et~al.(2020{\natexlab{b}})Zhang, Takanobu, Zhu, Huang, and
  Zhu}]{zhang2020recent}
Zheng Zhang, Ryuichi Takanobu, Qi~Zhu, MinLie Huang, and XiaoYan Zhu.
  2020{\natexlab{b}}.
\newblock Recent advances and challenges in task-oriented dialog systems.
\newblock \emph{Science China Technological Sciences}, pages 1--17.

\bibitem[{Zhao et~al.(2017)Zhao, Zhao, and Eskenazi}]{zhao2017learning}
Tiancheng Zhao, Ran Zhao, and Maxine Eskenazi. 2017.
\newblock Learning discourse-level diversity for neural dialog models using
  conditional variational autoencoders.
\newblock In \emph{Proceedings of the 55th Annual Meeting of the Association
  for Computational Linguistics (Volume 1: Long Papers)}, pages 654--664.

\bibitem[{Zhao et~al.(2022{\natexlab{a}})Zhao, Tian, Yao, Zheng, Lee, Song,
  Sun, and Zhang}]{zhao-etal-2022-improving}
Yingxiu Zhao, Zhiliang Tian, Huaxiu Yao, Yinhe Zheng, Dongkyu Lee, Yiping Song,
  Jian Sun, and Nevin Zhang. 2022{\natexlab{a}}.
\newblock \href {https://doi.org/10.18653/v1/2022.acl-long.44} {Improving
  meta-learning for low-resource text classification and generation via memory
  imitation}.
\newblock In \emph{Proceedings of the 60th Annual Meeting of the Association
  for Computational Linguistics (Volume 1: Long Papers)}, pages 583--595,
  Dublin, Ireland. Association for Computational Linguistics.

\bibitem[{Zhao et~al.(2022{\natexlab{b}})Zhao, Zheng, Yu, Tian, Lee, Sun, Li,
  and Zhang}]{zhao2022semi}
Yingxiu Zhao, Yinhe Zheng, Bowen Yu, Zhiliang Tian, Dongkyu Lee, Jian Sun,
  Yongbin Li, and Nevin~L. Zhang. 2022{\natexlab{b}}.
\newblock Semi-supervised lifelong language learning.
\newblock In \emph{Proceedings of the 2022 Conference on Empirical Methods in
  Natural Language Processing: Findings}.

\bibitem[{Zhu et~al.(2022)Zhu, Li, Mi, Zhu, and Huang}]{pmtdst}
Qi~Zhu, Bing Li, Fei Mi, Xiaoyan Zhu, and Minlie Huang. 2022.
\newblock Continual prompt tuning for dialog state tracking.

\end{thebibliography}
\bibliographystyle{acl_natbib}
\newpage
\appendix
\section{Details of Datasets}
\label{appendix:dataset}
We list the statistics of datasets for the intent detection and slot filling in Table ~\ref{tab:datasets} and give detailed descriptions as follows.
\begin{itemize}
    \item \textbf{ATIS} consists of audio recordings and corresponding manual transcripts about humans asking for flight information on automated airline travel inquiry systems. The data consists of 17 unique intent categories.  
    \item \textbf{BANKING} contains 13,083 utterances related to banking domain with 77 different fine-grained intents. 
    \item \textbf{CLINC} contains 10 domains (e.g., travel, kitchen, utility, etc.) and 150 different intent classes. 
    \item \textbf{DSTC} consists of slot annotations spanning 4 domains (buses, events, homes, rental cars).
    \item \textbf{HWU} includes 64 intents spanning 21 domains (e.g., alarm, music, IoT, news, calendar, etc.) 
    \item \textbf{MIT\_RESTAURANT} is a semantically tagged training and test corpus in BIO format.
    \item \textbf{MIT\_MOVIE} is a semantically tagged training and test corpus in BIO format. We choose ``eng'' corpus for implementation which consists of simple queries.
    \item \textbf{TOP} is a dataset of 44K utterances where each utterance is annotated with a hierarchical semantic representation.
    \item \textbf{SNIPS} contains crowdsourced queries distributed among 7 user intents of various complexity.
\end{itemize}

\begin{table}[thb]
\small
\begin{tabular}{lcccl}
\toprule
\multicolumn{5}{c}{Intent Detection Tasks}  \\
\midrule
Task & Train & Valid & Test & \# Intent \\
\midrule
ATIS    & 4.5K  & 0.5K  & 0.9K   & 17 \\
BANKING & 8.6K   & 1.5K  & 3.1K   & 77 \\
SNIPS   & 11.0K   & 1.3K  & 1.3K   & 7 \\
CLINC   & 15.0K   & 3.0K  & 4.5K   & 150 \\
HWU     & 8.9K  & 1.1K  & 1.1K   & 64 \\
TOP-S1  & 11.9K   & 1.7K  & 3.4K   & 6 \\
TOP-S2  & 11.9K   & 1.7K  & 3.4K   & 6  \\
TOP-S3  & 7.4K  & 1.0K  & 2.2K   & 7       \\
\midrule
\multicolumn{5}{c}{Slot Filling Tasks}  \\
\midrule
Task & Train & Valid & Test & \# Slot \\
\midrule
ATIS           & 4.5K & 0.5K & 0.8K  & 79 \\
SNIPS          & 11.0K  & 1.3K & 1.3K  & 39 \\
DSTC           & 3.7K & 1.8K & 1.8K  & 13 \\
MIT-MOVIE      & 8.5K & 1.2K & 2.4K  & 12\\
MIT-RESTAURANT & 6.1K & 1.5K & 1.5K  & 8 \\
\bottomrule
\end{tabular}
\caption{Statistics of datasets for intent detection and slot filling.}
\label{tab:datasets}
\end{table}

\section{Experiment Details}
\subsection{Prompt Examples of NLU Tasks}
\label{appendix:input}
We provide some detailed examples for inputs and outputs of the model with the designed prompts in PCLL.
For intent detection, when we train on ``BANKING'' task, an input utterance $x$ of the language model (LM) for a sample is modified as 
``\texttt{For an utterance from the BANKING task, ``I already have one of your cards, how do I link them?'' has the following intent }'',
the output of LM $y$ is its corresponding intent annotation: ``\texttt{Card linking}''.
For the ablation study of \textbf{w/o Task ID}, the prompt of the above sample becomes ``\texttt{For an utterance from the current task, ``I already have one of your cards, how do I link them?''}''.

For slot filling, when we train on the ``MIT-RESTAURANT'' task, an input utterance \texttt{x} is ``\texttt{Does the Casanova restaurant at Kendall Square offer a fixed price menu?}'' of LM is modified as ``\texttt{In the MIT-RESTAURANT task, if there are any slots and values, what are they in this sentence: ``Does the Casanova restaurant at Kendall Square offer a fixed price menu?''? Answer: }'', the output $y$ locating the contained slot-value pairs is modified as ``\texttt{Restaurant name: Casanova; Location: Kendall Square.}''. 
Here, different slot-value pairs are formatted as ``\texttt{slot: value}'' separated with ``;''. If the input $x$ does not contain any slot-value pairs, we use the sentence ``\texttt{No slot in this sentence.}'' as the output $y$.

\subsection{Different Task Orders}
\label{appendix:six_orders}
We list the six random permutations of tasks that we use to implement all competing methods in Table~\ref{tab:six_orders}.

\subsection{Model Implementation Details}
\label{appen:details}
We use a pre-trained GPT2 model \citep{gpt2} as the initialization for the encoder and decoder of CVAE in PCLL. 
We set the maximum context length as 256. Our model contains a total number of 240M parameters.
We train all competing methods on 1 Tesla-V100 GPU and it takes around 6 to 10 hours to train all the tasks. Moreover, the training and testing batch sizes are set to 64. The maximum learning rate is $5e-5$, the Adam optimizer is used with parameters $\beta_1=0.9$, $\beta_2=0.98$ and $\epsilon=1e-8$. 
The number of cycles for the cyclic annealing schedule is set to 4 in each epoch.
When generating pseudo samples, the maximum decoded sequence length is set to 96.  
For baselines implementations, we use BERT to implement HAT and CTR, and choose GPT-2 as the backbone model for other baselines (LAMOL, L2KD, ER, Adapter, EWC, MAS, Finetune).

\begin{table}[htbp]
    \centering
    \begin{tabular}{lcc}
    \toprule 
    & \multicolumn{1}{c}{Score} & LCA \\ 
    \midrule
    Prompt1 (12 tokens) & 90.25  & 88.82  \\
    Prompt2 (13 tokens) & 89.94  & 88.78  \\ 
    Prompt3 (4 tokens) & 90.34  & 88.75  \\
    Prompt4 (11 tokens) & 90.05  & 88.50  \\ 
    Prompt5 (28 tokens) & 89.20  & 88.22  \\
    \bottomrule
    \end{tabular}
    \caption{Applying different manual prompts on lifelong intent detection tasks. Each result is an average of 6 random task orders.}
    \label{tab:difpmts}
\end{table}

\subsection{Analysis of Manual Prompts Designs}
\label{appen:difpmt}
We list five different manual templates as the designed prompts of intent detection in Table~\ref{appen:multi_pmts}, where \texttt{Prompt1} is the one we use in Table~\ref{tab:main_res}. Let \texttt{ID} refers the task name, \texttt{x} refers the input utterance and \texttt{y} means the intent of \texttt{x}. 

\begin{table*}[t]
\centering
\small
\begin{tabular}{cl}
\toprule
& Different Manual Prompts for Intent Detection \\
\midrule
\texttt{Prompt1}& \texttt{For an utterance from the ID task, x has the following intent y} \\
\texttt{Prompt2}& \texttt{In the ID task, what intent best describes: x? Answer: y} \\
\texttt{Prompt3}& \texttt{Task ID utterance x intent y} \\
\texttt{Prompt4}& \texttt{In the task ID, this utterance x has the intent of y} \\
\texttt{Prompt5}& \texttt{If we consider the intent detection task, for a sample in the ID task, } \\
&  \texttt{what's the intent of the utterance x? The intent is: y} \\
\bottomrule
\end{tabular}
\caption{Different manual prompts are designed for intent detection module of a ToD system.}
\label{appen:multi_pmts}
\end{table*}

\subsection{Analysis of Prompt Attention}\label{appen:attn}
We provide the visualization of the attention scores over several samples employed with our designed prompts for intent detection tasks. Specifically, the attention score on each prompt token is calculated using the averaged attention it receives when generating the output prediction.
From the following Fig~\ref{fig:attn_case}, we can notice that the task names do contain meaningful information to be attended to when generating predictions.
\begin{figure*}[t]
    \centering
    \includegraphics[width=\textwidth]{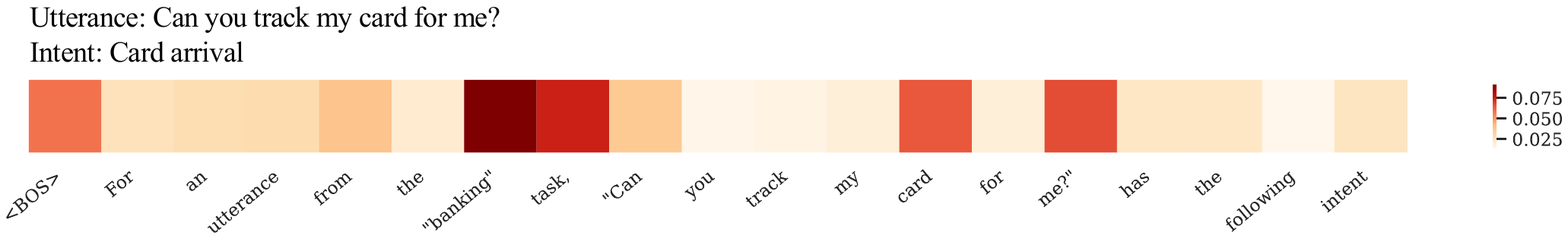}
    \caption{Visualization of attention scores for the natural language prompts of PCLL.}
    \label{fig:attn_case}
\end{figure*}

\subsection{Analysis of Pseudo-sample Quality}
\label{appen:pseudo-quality}
\begin{figure*}[ht]
    \centering
    \includegraphics[width=380px]{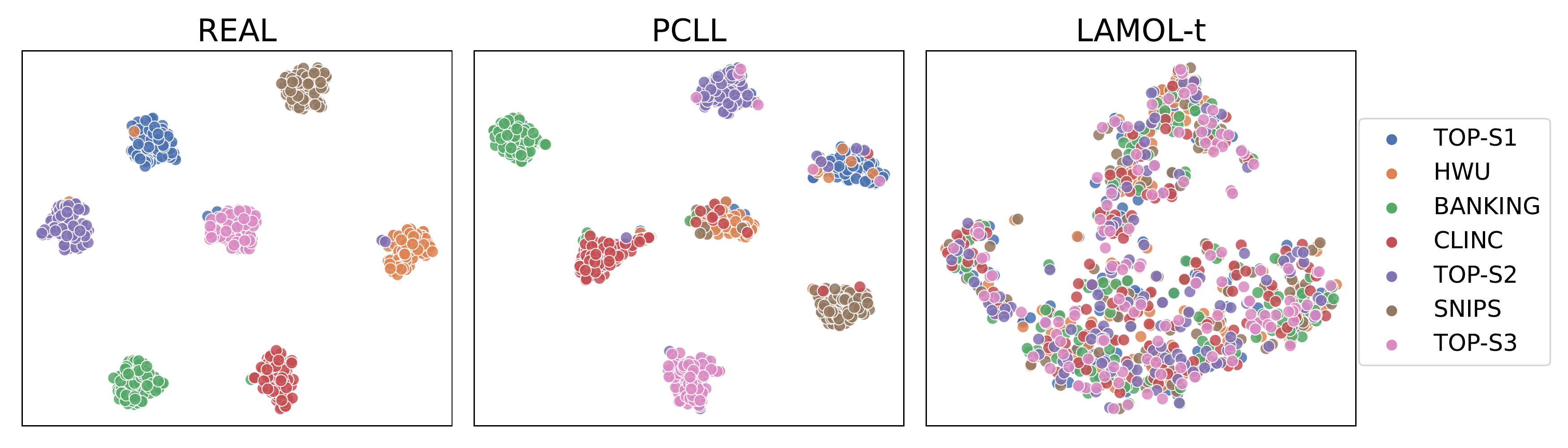}
    \caption{t-SNE visualization of the feature spaces associated with the generated pseudo samples.}
    \label{fig:tsne}
\end{figure*}
We analyze the quality of generated pseudo samples with PCLL and other generative replay-based baselines.
Specifically, we first fine-tune a pre-trained BERT \cite{devlin-etal-2019-bert} model using these observed real samples to construct a task classifier.
This classifier can determine the task identity of a given sample, and it reaches an accuracy of 98.67\% on a hold-out test set.
The fine-tuned BERT is used to extract the representation vector of each sample,
and the t-SNE algorithm \cite{tsne} is used to map these vectors into 2-dimensions.
For a specific task order \footnote{``TOP-S1, HWU, SNIPS, BANKDING, CLINC, TOP-S2, TOP-S3, ATIS''} in LL, we gather pseudo samples generated when learning the last task and visualize the feature space of these samples.
Note that the last task, ATIS, is not shown in Fig.~\ref{fig:tsne} since there is no need to replay the last task.
\begin{table*}[ht]
    \centering
    \scalebox{0.85}{
    \begin{tabular}{lc}
    \toprule
          & Intent Detection Tasks \\
         \midrule
         Order 1 & TOP\_S1, HWU, SNIPS, BANKING, CLINC, TOP\_S2, TOP\_S3, ATIS \\
         Order 2 & BANKING, HWU, TOP\_S1, TOP\_S3, CLINC, TOP\_S2, SNIPS, ATIS \\
         Order 3 & SNIPS, ATIS, TOP\_S2, TOP\_S3, CLINC, BANKING, HWU, TOP\_S1 \\
         Order 4 & CLINC, SNIPS, TOP\_S3, BANKING, TOP\_S2, HWU, TOP\_S1, ATIS \\
         Order 5 & BANKING, TOP\_S2, TOP\_S1, ATIS, TOP\_S3, HWU, CLINC, SNIPS \\
         Order 6 & CLINC, TOP\_S1, TOP\_S2, ATIS, SNIPS, HWU, BANKING, TOP\_S3 \\ 
         \midrule
         \midrule
         & Slot Filling Tasks \\ 
         \midrule
        Order 1 &  MIT\_MOVIE, DSTC, MIT\_RESTAURANT, SNIPS, ATIS \\
        Order 2 &  MIT\_MOVIE, SNIPS, DSTC, MIT\_RESTAURANT, ATIS \\
        Order 3 &  ATIS, MIT\_MOVIE, DSTC, MIT\_RESTAURANT, SNIPS \\
        Order 4 &  DSTC, MIT\_RESTAURANT, MIT\_MOVIE, ATIS, SNIPS \\
        Order 5 &  MIT\_MOVIE, ATIS, SNIPS, MIT\_RESTAURANT, DSTC \\
        Order 6 &  SNIPS, ATIS, MIT\_RESTAURANT, MIT\_MOVIE, DSTC \\
        \bottomrule
    \end{tabular} }
    \caption{Six random permutations of tasks for intent detection and slot filling.} 
    \label{tab:six_orders} 
\end{table*} 

\begin{table}[H]
\centering
\small
\begin{tabular}{lcc}
\toprule
\multicolumn{1}{c}{Ratio $\gamma$} & \multicolumn{1}{c}{Score} & LCA \\ 
\midrule
0.01             & \multicolumn{1}{c}{73.61}    & 84.35     \\
0.05             & \multicolumn{1}{c}{84.09}    & 89.54     \\ 
0.20              & \multicolumn{1}{c}{90.25}   & 88.82     \\ 
0.50              & \multicolumn{1}{c}{91.02}   & 91.44     \\  
1.00              & \multicolumn{1}{c}{91.31}   & 91.77     \\ 
\bottomrule
\end{tabular}
\caption{The LL performance on various sampling ratio $\gamma$. Each result is an average of 6 random task orders.}
\label{tab:ratio}
\end{table}

\subsection{Analysis of Sampling Ratio}
\label{appen:sample_ratio}
Table \ref{tab:ratio} shows the results on intent detection tasks.
It can be seen that generating more pseudo samples helps to improve the LL performance.
Besides, the performance gain slows down as the sampling ratio $\gamma$ exceeds $0.2$,
i.e., generating 5 times more pseudo samples from $\gamma = 0.01$ to $\gamma = 0.05$ yields 10.48 absolute improvement on the \textit{Score} metric,
while increasing $\gamma$ from 0.2 to 1.0 only yields 1.63 absolute improvement.

\section{Case Study}\label{appen:case}
We present more generated pseudo samples from PCLL and LAMOL along with real samples in Table~\ref{tab:more_cases}.   
For intent detection, we list real and pseudo samples from HWU tasks; for slot filling, we list those samples from MIT-RESTAURANT and DSTC tasks in Table~\ref{tab:more_cases}. 

\begin{table*}[t]
\centering
\scalebox{0.65}{
\begin{tabular}{lll}
\toprule
                                           & Utterance                                                                                              & Labels                                              \\
\midrule                                           
\multicolumn{3}{c}{Intent Detection (HWU)}                                                                                                                                                                \\
\midrule
\multirow{3}{*}{REAL}                      & What is the most current exchange rate in china.                         & qa\_currency                            \\
                                           & Book a train ticket to Canada this evening.                              & transport\_ticket                  \\
                                           & Set up each day an alarm at ten time for my meal.                        & alarm set                               \\
\midrule                                           
\multirow{3}{*}{PCLL}                      & How many currencies can I exchange for a one pound dollar?               & qa\_currency                          \\
                                           & Book a train ticket to make it home by five pm.                          & transport\_ticket                       \\
                                           & Please set a new alarm for nine am on Friday.                              & alarm set                          \\
\midrule                                           
\multirow{3}{*}{LAMOL}                     & How much money is going on for the election.                             & qa\_currency                    \\
                                           & I want to stay in the next week and get a decent.                        & GET\_EVENT                        \\
                                           & I am home from work tomorrow evening.                                    & alarm set              \\
\midrule                                           
\multicolumn{3}{c}{Slot   Filling (MIT-RESTAURANT)}                                                                                                                                                                 \\
\midrule
\multirow{3}{*}{REAL}                      & Does the caranova restaurant at kendall square offer a fixed price menu.                              & Restaurant name: caranova; Location: kendall square \\
                                           & Could you find me a place that's open every day.                                                       & Hours: open every day                               \\
                                           & I'm in the mood for chinese food.                                                                        & Cuisine: Chinese                                    \\
\midrule                                           
\multirow{3}{*}{PCLL}                      & What pizza place that serves pizza in Manhattan has a great prices and is open at 2 am.               & Cuisine: pizza; Location: manhattan                 \\
                                           & What restaurant has good beer that is open before 9 am.                                                 & Hours: open before 9 am                             \\
                                           & Find me a vegetarian restaurant with outdoor seating.                                                   & Cuisine: vegetarian; Amenity: outdoor seating       \\
\midrule                                           
\multirow{3}{*}{LAMOL}                     & I would like to rent a room in this area.                                                            & area: this area                                     \\
                                           & Does this place offer an open menu.                                                                   & Hours: open menu                                    \\
                                           & The reason: It has been a while.                                                                      & No slot in this sentence.             \\
\midrule                                           
\multicolumn{3}{c}{Slot Filling (DSTC)}                                                                                                                                                        \\
\midrule
\multirow{3}{*}{REAL}                      & Would I would be able to get another car for the 10th of this month?                                 & dropoff\_date: 10th of the month                    \\
                                           & I plan to go to Hayward.                                                                               & area: Hayward                                       \\
                                           & I need to pick it up from SD next Thursday.                                                          & Pickup\_city: SD; Pickup\_data: next Thursday       \\
\midrule                                           
\multirow{3}{*}{PCLL}                      & I need the car until 11th of March, can you find me the available one.                                & dropoff\_date: 11th of March                        \\
                                           & The one in San Mateo is not fully furnished.                                                           & area: San Mateo                                     \\
                                           & Thanks for not leaving on 2nd of March and I would be in Long Beach, CA.                               & pickup\_city: Long Beach, CA                        \\
\midrule                                           
\multirow{3}{*}{LAMOL}                     & I am looking for an apartment in San Francisco, CA.                                                   & area: San Francisco, CA                             \\
                                           & I want to purchase the car, and it's parked on the street.                                            & dropoff\_date: on the street                       \\
                                           & I need to get to the city.                                                                             & pickup\_city: Seattle    \\                          
\bottomrule                                           
\end{tabular}}
\caption{Real samples and generated pseudo samples by PCLL and LAMOL-t.}
\label{tab:more_cases}
\end{table*}

\section{Analyses of Forgetting}
\label{appen:forgetting}
We provide more ﬁne-grained analyses for the forgetting issue.
\begin{itemize}[leftmargin=*]
\item Classes with fewer samples are easier to be forgotten. Some tasks (e.g., ATIS, TOP, MIT-MOVIE) have unbalanced classes. These minor classes that only occupy a small portion of training samples are less likely to appear in pseudo samples used for replay. For example, the intent “meal” only takes 0.13\% of the training samples for ATIS, and there are barely any pseudo samples generated for this intent when replaying. Without these pseudo samples, the model is more likely to forget these minor classes.
\item Different tasks may have partially overlapping data distributions and conﬂicted label spaces, i.e., some tasks may assign different labels to the same set of utterances. For example, in the CLINC dataset, the utterance “transfer funds to the other account” is assigned with a label of “transfer”; however, in the BANKING dataset, the same utterance is assigned with a label of “transfer into account”. These conﬂicted label spaces may confuse the model, resulting in incorrectly labeled pseudo samples.

\item Noisy pseudo labels created by generative replay may lead to error accumulation, which will downgrade the performances of previously learned tasks.
\item The diversity of generated pseudo samples for previous tasks tends to decrease as more replay times are performed, and these less diversiﬁed pseudo samples lead to more forgetting. 

Speciﬁcally, we conduct analyses on lifelong intent detection tasks with the following task order (CLINC, SNIPS, TOP\_S3, BANKING, TOP\_S2, HWU, TOP\_S1, ATIS). We compare the diversity of pseudo-samples for the ﬁrst task (i.e., CLINC) generated at different replay moments: (1) after learning the ﬁrst task, (2) after learning three subsequent tasks, and (3) after learning eight subsequent tasks (i.e., after the last task’s learning). 

In Table~\ref{tab:afterdist}, we use the distinct scores \citep{dist} to measure the diversity of pseudo samples. We can notice that as we learn more tasks, the diversity of pseudo samples for the ﬁrst learned task decreases. Therefore, replaying less diverse pseudo samples leads to performance degradation on previous tasks (i.e., forgetting of previous tasks).
\end{itemize}

\begin{table}[H]
\centering
\scalebox{0.85}{
\begin{tabular}{ccccc}
\toprule
After N Tasks & Dist-1 & Dist-2 & Dist-3 & Dist-4 \\
\midrule
1                      & 0.3593 & 0.7985 & 0.9439 & 0.9838 \\
4                      & 0.3193 & 0.7308 & 0.8951 & 0.9526 \\
8                      & 0.3091 & 0.6927 & 0.8593 & 0.9301 \\
\bottomrule
\end{tabular}}
\caption{Diversity scores of generated pseudo samples after learning N tasks.}
\label{tab:afterdist}
\end{table}

\end{document}